\documentclass[graybox]{svmult}

\usepackage{float}

\usepackage{type1cm}        

\usepackage{makeidx}         
\usepackage{graphicx}        
\usepackage{multicol}        
\usepackage[bottom]{footmisc}

\usepackage{xspace} 
\usepackage{newtxtext}       
\usepackage{newtxmath}       
\usepackage{afterpage} 
\newcommand{\MATLAB}{\textsc{Matlab}\xspace}

\makeindex            	

\usepackage{amsmath, epsfig}

\usepackage{mathrsfs}
\usepackage{color}
\usepackage{ifthen}
\usepackage{tikz}
\usetikzlibrary{arrows}
\usepackage{comment}

\usepackage{algorithmicx}
\usepackage[ruled]{algorithm}
\usepackage{algpseudocode}
\usepackage{graphicx}
\usepackage{caption}
\usepackage{subcaption}
\usepackage[mathscr]{euscript}
\usepackage{enumerate}

\usepackage[font=small,margin=10pt,labelfont={bf},labelsep={space}]{caption}

\usepackage[scaled]{helvet} 
\usepackage{bm} 
\usepackage{fullpage}

\usepackage{hyperref}
\theoremstyle{plain}

\numberwithin{theorem}{section}
\numberwithin{equation}{section}

\DeclareMathOperator*{\argmin}{arg min}

\def \Re {\mathbb{R}}

\def \E {\mathbb{E}}

\def \R {\mathbb{R}}

\def \rank {{\rm rank }}

\begin{document}

\title*{On Large-Scale Dynamic Topic Modeling with Nonnegative CP Tensor Decomposition\thanks{The initial research for this effort was conducted at the Research Collaboration Workshop for Women in Data Science and Mathematics, July 2019 held at ICERM. Funding for the workshop was provided by ICERM, AWM and DIMACS (NSF grant CCF1144502).  Haddock and Needell were partially supported by NSF CAREER $\#1348721$ and NSF BIGDATA $\#1740325$.}}
\titlerunning{On Large-Scale Dynamic Topic Modeling}

\author{Miju Ahn, Nicole Eikmeier, Jamie Haddock, Lara Kassab, Alona Kryshchenko, Kathryn Leonard, Deanna Needell, R. W. M. A. Madushani, Elena Sizikova, Chuntian Wang}

\institute{
Miju Ahn \at Department of Engineering Management, Information, and Systems, Southern Methodist University, Dallas, TX, 75205, U.S.A, \email{mijua@smu.edu}
\and 
Nicole Eikmeier \at Department of Computer Science, Grinnell College,  Grinnell, IA, 50112, U.S.A, \email{eikmeier@grinnell.edu}
\and
Jamie Haddock \at Department of Mathematics, University of California, Los Angeles, CA, 90095-1555, U.S.A, \email{jhaddock@math.ucla.edu}
\and 
Lara Kassab \at Department of Mathematics, Colorado State University, Fort Collins, CO, 80523, U.S.A, \email{kassab@math.colostate.edu}
\and
Alona Kryshchenko \at Department of Mathematics, California State University, Channel Islands, Camarillo, CA, 93012, U.S.A, \email{alona.kryshchenko@csuci.edu}
\and
Kathryn Leonard \at Department of Computer Science, Occidental College, Los Angeles, CA, 90041, U.S.A, \email{kleonard.ci@gmail.com}
\and
Deanna Needell \at Department of Mathematics, University of California, Los Angeles, CA, 90095, U.S.A, \email{deanna@math.ucla.edu}
\and
R. W. M. A. Madushani \at Section of Infectious Diseases, Boston Medical Center, MA, 02118, U.S.A,
\email{madushani.rajapaksha@bmc.org}
\and
Elena Sizikova \at 
Center for Data Science, New York University, New York, NY, 10011, U.S.A 
\email{es5223@nyu.edu}
\and
Chuntian Wang \at The Department of Mathematics, University of Alabama, Tuscaloosa, AL, 35487, U.S.A
\email{cwang27@ua.edu}
}

\maketitle

\vspace{-1in}

\abstract{There is currently an unprecedented demand for large-scale temporal data analysis due to the explosive growth of data. Dynamic topic modeling has been widely used in social and data sciences with the goal of     learning latent topics that emerge, evolve, and fade over time. Previous work on dynamic topic modeling primarily employ the method of nonnegative matrix factorization (NMF), where slices of the data tensor are each factorized into the product of lower-dimensional nonnegative matrices. With this approach, however, information contained in the temporal dimension of the data is often neglected or underutilized. To overcome this issue, we propose instead adopting the method of nonnegative CANDECOMP/PARAPAC (CP) tensor decomposition (NNCPD), where the data tensor is directly decomposed into a minimal sum of outer products of nonnegative vectors, thereby preserving the temporal information. The viability of NNCPD is demonstrated through application to both synthetic and real data, where significantly improved results are obtained compared to those of typical NMF-based methods. The advantages of NNCPD over such approaches are studied and discussed. To the best of our knowledge, this is the first time that NNCPD has been utilized  for the purpose of dynamic topic modeling, and our findings will be transformative for both applications and further developments. }

\section{Introduction}

In today's society, there is an unprecedented
demand for efficient, quantitative, and interpretable methods to study large-scale data in various fields such as finance, economy, social media, psychology, and political sciences (see, e.g., \cite{doyle2009financial,ambrosino2018topic,saha2012learning,bittermann2018identify,yano2009predicting,cichocki2007nonnegative, chen2019modeling, bahargam2018constrained}. The area of study on which we focus is known as \emph{topic modeling}, which investigates ways to reveal latent themes and topics in a dataset. \emph{Dynamic topic modeling} (see, e.g., \cite{cichocki2007nonnegative,traore2018non,saha2012learning}) investigates how topics emerge (i.e., new topics are formed), evolve (i.e., topics gradually change meaning), and fade (i.e., topics disappear in importance), and is of particular interest for the analysis of data with a temporal component.

Data obtained in the applications mentioned above is often of high dimension, including one or more temporal dimensions, and is well-represented by tensors, a common algebraic representation for high-dimensional arrays (see \cite{kolda2009tensor} for a tutorial).  The crucial step of (dynamic) topic modeling is to decompose high-dimensional tensors into interpretable representations with attention to the temporal information. 
In addition, one may also be interested in finding such decompositions with some additional structure, such as nonnegativity, which allows for interpretability of topics as opposed to traditional approaches like principal component analysis (PCA) where factors often cancel.  

In previous works, the typical methods for such nonnegative tensor decompositions are mainly based on  nonnegative matrix factorization (NMF) where a \emph{matricized} version of the tensor sliced along the temporal dimension is factorized using NMF. There are two basic approaches for the NMF-based nonnegative tensor decomposition: (i) \emph{NMF applied directly to  tensor slices (Direct NMF)}, where a tensor is broken into slices along the temporal dimension and each time slice is decomposed independently using  
NMF \cite{lawton1971self,paatero1994positive,anttila1995source,lee1999learning}; and (ii) \emph{Fixed-factor Nonnegative Matrix Factorization (Fixed NMF)}, where the slices along the temporal dimension are concatenated together and decomposed with one of the factors being fixed~\cite{cichocki2007nonnegative}. More advanced NMF-based approaches based on these two basic ones have been developed, e.g., a windowing technique, where  multiple temporal slices are considered at once, which forces factorizations of nearby slices in the temporal dimension to be similar  \cite{saha2012learning, chen2015modeling}.  

A significant drawback of NMF-based nonnegative tensor decompositions, however, is their failure to respect  the temporal mode. For Direct NMF, the data are treated as ``independent'' across time. For Fixed NMF, the data are assumed to share the same latent topics over time. Neither assumption holds true in many application domains.  Moreover, it is often such \textit{changes} in the topic structure and information that is relevant to the application, and it is those changes that need to be identified. Therefore, it is imperative to find a tensor decomposition method  that captures the full information of the  topics evolving over time. To  this end, we propose to adopt the Nonnegative CANDECOMP/PARAPAC tensor decomposition (NNCPD)~\cite{carroll1970analysis,harshman1970foundations} for dynamic topic modeling, where the tensor is directly decomposed into sums of outer products of (one-dimensional) vectors.  In \cite{chen2019modeling, bahargam2018constrained}      coupled tensor-matrix  factorization models are adopted.  The nonnegative tensor decomposition is coupled   with     observable auxiliary matrices providing  information for the decomposition. Here, our proposed NNCPD is  unsupervised,     which is more suitable than  semi-supervised   methods   when  there is no a prior knowledge about the tensor decomposition or structure. Although there have been many applications of NNCPD to various areas of data analysis \cite{kolda2009tensor}, to the best of our knowledge, it has not been applied to dynamic topic modeling blue
{in an unsupervised format}.

{\bfseries Contribution.}  To compare the method  of NNCPD to the traditional NMF-based methods, we consider dynamic topic modeling of synthetically generated datasets, as well as those drawn from the 20 Newsgroup Data \cite{Lang95,20NewsGroup} which is a large-scale benchmark dataset used in testing topic modeling methods. The numerical experiments demonstrate that the NNCPD outperforms both Direct NMF and Fixed NMF in two key ways:
\begin{enumerate}
\item  NNCPD incorporates the temporal dimension in its factors, regardless of which mode captures temporal information, and detects the dynamics of the topics through time, i.e., the emergence, evolution, and fading of topics. These phenomena are clearly visualized by NNCPD, while the other methods may obscure these phenomena.

\item   NNCPD is more robust to the approximation of the number of topics and noise in the data than the other methods. When the true number of topics may be slightly overestimated, the other methods often fit topics to the noise. 
In contrast, 
the performance of NNCPD does not significantly deteriorate for data with a reasonable signal-to-noise ratio. 
\end{enumerate}

{\bfseries Organization.} The paper is organized as follows. In Sect. \ref{sec:overview}, we give a brief overview on NMF and the tensor decompositions considered in the article.
We then illustrate NNCPD as a tool for topic detection for dynamic topic modeling on three-way synthetic data (Sect. \ref{sec:syntheticdataset}) and   the 20 Newsgroup data  (Sect. \ref{sec:20news}), and perform numerical experiments to test robustness of NNCPD (Sect.~\ref{sec: robustness}). The output of our NNCPD experiments are compared to those of Direct NMF and Fixed NMF, and the advantages of NNCPD over these typical NMF-based methods are analyzed and discussed. We conclude with discussion and remarks in Sect. ~\ref{sec:conclusion}.

\section{Overview and Notations}
\label{sec:overview}

In this section, we introduce the basic notions of NMF-based nonnegative tensor decompositions and NNCPD. Formally, a \textit{tensor} is a multidimensional array and the \textit{order} of a tensor is the number of dimensions, also known as ways or modes. A first-order tensor is a vector, a second-order tensor is a matrix, and tensors of order three or higher are called higher-order tensors~\cite{kolda2009tensor}. In what follows, to distinguish between decompositions for tensors and matrices,
we will use the term \emph{factorization} only in reference to matrix factorization (including matrix factorization subroutines in some tensor analysis methods) and the term \emph{decomposition} only in reference to tensor decomposition.

\subsection{NMF-based Nonnegative Tensor Decompositions}
\label{sec:nmf for tensors}

In this section, we introduce two typical NMF-based nonnegative tensor decompositions: Direct NMF~\cite{lawton1971self,paatero1994positive,anttila1995source,lee1999learning} and Fixed NMF~\cite{cichocki2007nonnegative}.  We start with a general introduction of NMF for matrices. 

\subsubsection{NMF for Matrices}
\label{sec:nmf description}

Nonnegative matrix factorization (NMF) seeks to find an approximate factorization of a nonnegative data matrix $X \in \mathbb{R}^{n_1 \times n_2}_{\ge 0}$ into a nonnegative features matrix $A$ and a nonnegative coefficients matrix $B$
\begin{equation}\label{eq:NMF}
X \approx AB, \quad A \in \mathbb{R}^{n_1 \times r}_{\ge 0}, B \in \mathbb{R}^{r \times n_2}_{\ge 0},
\end{equation}
where $r \in \mathbb N$ corresponds to the number of latent \textit{topics} in the data, and is typically much smaller than $n_1$ and $n_2$. We note that the outer product representation of matrix multiplication lets us rewrite the product $AB$ as 
\begin{equation}\label{eq:nmfproduct}
X\approx AB = \sum_{l=1}^r a_l \otimes b_l,
\end{equation}where $a_l \in \mathbb{R}^{n_1}_{\ge 0}$ is a column of $A$ and $b_l \in \mathbb{R}^{n_2}_{\ge 0}$ is a row of $B$.  
Generally, the factorization is computed by approximately minimizing the reconstruction error 
\begin{equation}\label{eq:energy}
    E (X; A, B) =   \| X - AB \|_F,
\end{equation}
where $\| \cdot \|_F$ denotes the Frobenius norm  of matrices. When the minimum of reconstruction error vanishes we say an exact NMF  is obtained. 

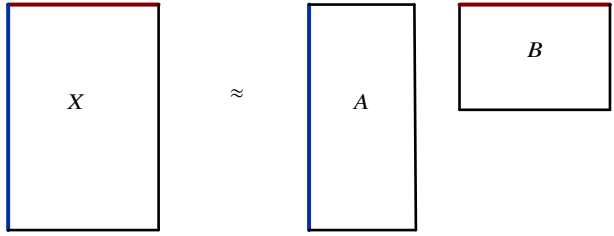
\begin{figure}[h!]
	\centering
		\definecolor{qqttzz}{rgb}{0.,0.2,0.6}
	\definecolor{yqqqqq}{rgb}{0.5019607843137255,0.,0.}
	\definecolor{qqwwtt}{rgb}{0.,0.4,0.2}
	\begin{tikzpicture}[line cap=round,line join=round,x=10.0cm,y=10.0cm,scale=0.1]
	\clip(-4.084355522389028,-0.142937766044919) rectangle (4.029312599052434,3.04887981977488);
	\draw [line width=1.5pt,color=yqqqqq] (-4.,3.)-- (-2.,3.);
	\draw [line width=1.5pt,color=qqttzz] (-4.,3.)-- (-4.,0.);
	\draw [line width=1.pt] (-4.,0.)-- (-2.,0.);
	\draw [line width=1.pt] (-2.,3.)-- (-2.,0.);
	\draw [line width=1.5pt,color=qqttzz] (0.,3.)-- (0.,0.);
	\draw [line width=1.pt] (0.,0.)-- (1.42,0.);
	\draw [line width=1.pt] (1.42,0.)-- (1.4,3.);
	\draw [line width=1.pt] (0.,3.)-- (1.4,3.);
	\draw [line width=1.pt] (2.,3.)-- (2.,1.6);
	\draw [line width=1.5pt,color=yqqqqq] (2.,3.)-- (4.,3.);
	\draw [line width=1.pt] (2.,1.6)-- (4.,1.6);
	\draw [line width=1.pt] (4.,1.6)-- (4.,3.);
	\draw (-1.1601151097619953,1.9908394864992116) node[anchor=north west] {$\approx$};
	\draw (-3.3250986348610972,1.9137028052847974) node[anchor=north west] {$X$};
	\draw (0.4642549973623304,1.9137028052847974) node[anchor=north west] {$A$};
	\draw (2.7809285184938643,2.6022495301424536) node[anchor=north west] {$B$};
	\end{tikzpicture}
	\caption{A visualization of the factor matrices in NMF of $X\approx A B$, where   $X \in \mathbb{R}^{n_1 \times n_2}_{\ge 0}$, $A \in \mathbb{R}^{n_1 \times r}_{\ge 0}$ and $B \in \mathbb{R}^{r \times n_2}_{\ge 0}$. The edges of the matrix visualized in blue and red represent the modes of the matrix with dimension $n_1$ and $n_2$, respectively.}\label{fig:nmf_figure}
\end{figure}

In an application where $X$ is a data matrix whose columns represent documents and whose rows represent words, the matrix $A$ gives a topic representation for each word, and $B$ a topic representation for each document. Furthermore, one might hope that the topics detected by NMF truly correspond to a set of topics that describe the set of documents well, and this is often the case in application of NMF to data.  
See Figure~\ref{fig:nmf_figure} for a visualization of NMF as in (\ref{eq:NMF}). 

\subsubsection{Direct NMF and Fixed NMF}

Datasets considered in the context of dynamic modeling are often represented as  higher-order tensors, for example, a data tensor  $X \in \Re_{\geq 0}^{n_1 \times n_2 \times n_3}$, whose first, second, and third modes represent documents, words, and time, respectively. A natural   way to decompose this third-order tensor is to perform NMF on temporal mode slices (or collections thereof)  of the tensor. There are two basic approaches of NMF-based nonnegative tensor decomposition: Direct NMF and Fixed NMF. 

Direct NMF on tensor slices performs NMF independently on each slice of the tensor~\cite{lawton1971self,paatero1994positive,anttila1995source,lee1999learning}. Given $X  \in \Re_{\geq 0}^{n_1 \times n_2 \times n_3}$, slicing along the third mode gives nonnegative matrices $X_i \in  \Re_{\geq 0}^{n_1 \times n_2 }$ for $i = 1, 2, \cdots, n_3$, each of which is factored into nonnegative matrices
\begin{equation}\label{eq:directnmf}
X_i \approx A_i S_i, \quad  \quad A_i \in \mathbb{R}^{n_1 \times r}_{\ge 0}, S_i \in \mathbb{R}^{r \times n_2}_{\ge 0}, \quad  i = 1, ..., n_3,
\end{equation}
where the $A_i$'s will be referred to as the Direct NMF $A$ factors, and the $S_i$'s the Direct NMF $S$ factors. This form of nonnegative tensor decomposition fails to capture inherent structures within the tensor along the time dimension. 
{Stacking the products of the Direct NMF $A$ and $S$ matrices forms an approximation to $X$, which will be referred to as the Direct NMF reconstruction.} The reconstruction error for Direct NMF 
 is defined  as:
\begin{equation}\label{eq:energy2}
    E (X; A_i, S_i, i=1, ..., n_3) = 
    \| X - \hat X \|_F,
\end{equation}
where $\hat X$ denotes the Direct NMF reconstruction. 
In the subsequent numerical experiments, the reconstruction error will be measured either with the data tensor or the data tensor corrupted with noise perturbation; the specific tensor used will be indicated in each situation.

In~\cite{cichocki2007nonnegative}, the authors define an alternative nonnegative tensor decomposition, which we refer to as \emph{Fixed NMF}.  This decomposition performs NMF simultaneously on the $n_3$ slices along mode three,  $X_i$, $i=1, ..., n_3$, with the same $A$. They consider a sequence of nonnegative matrix factorizations $(A,S_1), \cdots , (A,S_{n_3})$ such that 
\begin{equation}\label{eq:fnmf}
X_i \approx A S_i, \quad  \quad A  \in \mathbb{R}^{n_1 \times r}_{\ge 0}, S_i \in \mathbb{R}^{r \times n_2}_{\ge 0}, \quad  i = 1, ..., n_3,
\end{equation}
where  $A$ will be referred to as the Fixed NMF common $A$ factor, and the $S_i$'s the Fixed NMF $S$ factors. In other words, Fixed NMF fixes a single dictionary matrix $A$ and searches for the representations $S_i$ for each of the slices $X_i$. 
{Stacking the products of the Fixed NMF $A$ matrix and $S$ matrices forms an approximation to $X$, which will be referred to as the Fixed NMF reconstruction.}
The Fixed NMF reconstruction error is defined as:
\begin{equation}\label{eq:rec_error_fixed_nmf}
    E (X; A, S_i, i=1, ..., n_3) = 
    \| X - \hat X \|_F,
\end{equation}
where $\hat X$ denotes the  the Fixed NMF reconstruction. In the subsequent numerical experiments, the reconstruction error will be measured either with the data tensor or the data tensor corrupted with noise perturbation; the specific tensor used will be indicated in each situation.

\subsection{CANDECOMP/PARAFAC (CP) Decomposition and NNCPD}

In this section, we introduce the nonnegative CANDECOMP/PARAFAC (CP) decomposition (NNCPD), which generalizes NMF for matrices to higher-order tensors \cite{carroll1970analysis,harshman1970foundations}.

\subsubsection{Methodology of CP Decomposition and NNCPD}
Unlike the NMF-based nonnegative tensor decompositions, CP decompositions treat the tensor as a whole. The CP decomposition and NNCPD factorize  a tensor into a sum of component rank-one
tensors without slicing it along the temporal mode. For example, given a third-order tensor $X \in \R^{n_1 \times n_2 \times n_3}$, an exact rank-$r$ CP decomposition of $X$ can be written as
\begin{equation}\label{eq:cpd}
X = \sum \limits_{  \ell= 1}^r a_\ell \otimes b_\ell \otimes c_\ell, 
\end{equation}
where $\otimes$ denotes the outer product and $a_\ell \in \Re ^{n_1}, b_\ell \in \Re ^{n_2}, c_\ell \in \Re ^{n_3}$, $\ell = 1, ..., r$. 
Further, we can explicitly write out the entries of $X$ as follows, 
\begin{equation} 
x_{ijk} = \sum \limits_{\ell = 1}^r  a_{\ell i} b_{\ell j}  c_{\ell k},
\label{eq:CP decomp} 
\end{equation}
where $i = 1, \cdots n_1$, $j = 1, \cdots n_2$, and $k = 1, \cdots n_3$. The \emph{factor matrices} of  CP decomposition refer to the combination of the vectors from the rank-one components, i.e., $A = [a_1 \quad a_2 \quad \cdots \quad a_r]$, and likewise for $B$ and $C$.   The rank of the tensor $X$, denoted $\rank(X)$, is the smallest integer $r$ so that $X$ may be expressed as the sum of exactly $r$ rank-one tensors. Similar to NMF, an approximate CP decomposition may be computed. Fix an $r$ and approximately minimize the reconstruction error
\begin{equation}\label{eq:tensorerror}
    E(X; A, B, C) = \| X - \hat X \|_F.
\end{equation}
The solution $\hat X =\sum \limits_{  \ell= 1}^r a_\ell \otimes b_\ell \otimes c_\ell $ will be referred to as a rank-$r$ CP reconstruction   of $X$. In the subsequent numerical experiments, the reconstruction error will be measured either with the data tensor or the data tensor corrupted with noise perturbation; the specific tensor used will be indicated in each situation. In Sect.~\ref{sec: robustness}, we define the reconstruction error as $\| X-\hat T\|_F$ where $X$ is the original tensor and $\hat T$ is the reconstruction of the noisy tensor $T$ using the associated method.
This notion of reconstruction error will help us in understanding NNCPD as a denoising method, as we do not wish to fit the noise $N$.

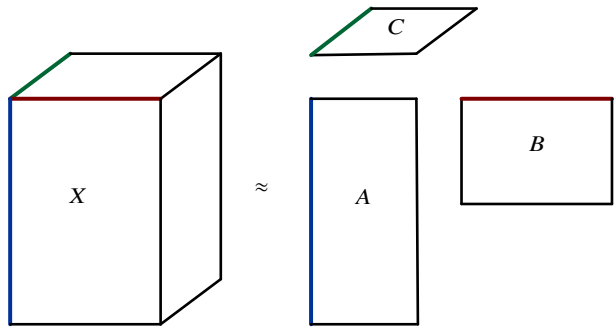
\begin{figure}[h!]
	\centering
		\definecolor{qqttzz}{rgb}{0.,0.2,0.6}
	\definecolor{yqqqqq}{rgb}{0.5019607843137255,0.,0.}
	\definecolor{qqwwtt}{rgb}{0.,0.4,0.2}
	\begin{tikzpicture}[line cap=round,line join=round,x=10.0cm,y=10.0cm,scale=0.1]
	\clip(-4.084355522389028,-0.142937766044919) rectangle (4.029312599052434,4.24887981977488);
	\draw [line width=1.5pt,color=qqwwtt] (-4.,3.)-- (-3.2,3.6);
	\draw [line width=1.pt] (-3.2,3.6)-- (-1.2,3.6);
	\draw [line width=1.pt] (-1.2,3.6)-- (-2.,3.);
	\draw [line width=1.5pt,color=yqqqqq] (-4.,3.)-- (-2.,3.);
	\draw [line width=1.5pt,color=qqttzz] (-4.,3.)-- (-4.,0.);
	\draw [line width=1.pt] (-4.,0.)-- (-2.,0.);
	\draw [line width=1.pt] (-2.,3.)-- (-2.,0.);
	\draw [line width=1.pt] (-1.2,3.6)-- (-1.2,0.6);
	\draw [line width=1.pt] (-2.,0.)-- (-1.2,0.6);
	\draw [line width=1.5pt,color=qqttzz] (0.,3.)-- (0.,0.);
	\draw [line width=1.pt] (0.,0.)-- (1.42,0.);
	\draw [line width=1.pt] (1.42,0.)-- (1.4,3.);
	\draw [line width=1.pt] (0.,3.)-- (1.4,3.);
	\draw [line width=1.pt] (2.,3.)-- (2.,1.6);
	\draw [line width=1.5pt,color=yqqqqq] (2.,3.)-- (4.,3.);
	\draw [line width=1.pt] (2.,1.6)-- (4.,1.6);
	\draw [line width=1.pt] (4.,1.6)-- (4.,3.);
	\draw [line width=1.pt] (0.,3.58)-- (1.4,3.6);
	\draw [line width=1.5pt,color=qqwwtt] (0.,3.58)-- (0.8,4.2);
	\draw [line width=1.pt] (0.8,4.2)-- (2.2,4.2);
	\draw [line width=1.pt] (2.2,4.2)-- (1.4,3.6);
	\draw (-0.8601151097619953,1.9908394864992116) node[anchor=north west] {$\approx$};
	\draw (-3.3250986348610972,1.9137028052847974) node[anchor=north west] {$X$};
	\draw (0.4642549973623304,1.9137028052847974) node[anchor=north west] {$A$};
	\draw (2.7809285184938643,2.6022495301424536) node[anchor=north west] {$B$};
	\draw (0.9085283597911595,4.16647966107218) node[anchor=north west] {$C$};
	\end{tikzpicture}
	\caption{A visualization of the factor matrices in a CP decomposition. The edges of the tensor visualized in blue, red, and green represent the modes of the tensor with dimension $n_1$,$n_2$, and $n_3$, respectively.}\label{fig:NNCP}
\end{figure}

When the reconstruction error vanishes, an exact CP decomposition as in (\ref{eq:cpd}) is obtained.  
See Figure~\ref{fig:NNCP} for a visualization of the rank-$r$ CP decomposition.

Note that NMF specializes matrix factorization to factorizing a nonnegative data matrix into the product of two (lower-dimensional) nonnegative factor matrices.  In the same way, NNCPD specializes the CP decomposition to decomposing a nonnegative data tensor into the sum of rank-one tensors which are the outer product of nonnegative vectors. Nonnegativity is necessary when we desire to preserve  inherent properties of the original tensor data. For example, a tensor of images will have entries representing pixel values that must be nonnegative. 
Specifically, given a third-order tensor $X \in \R^{n_1 \times n_2 \times n_3}_{\ge 0}$ and a fixed integer $r$, the approximate NNCPD of $X$ seeks $A \in \mathbb{R}^{n_1 \times r}_{\ge 0}, B \in \mathbb{R}^{n_2 \times r}_{\ge 0}, C \in \mathbb{R}^{n_3 \times r}_{\ge 0}$ so that
\begin{equation}\label{eq:nncpd}
  X \approx \sum \limits_{  \ell= 1}^{r} a_\ell \otimes b_\ell \otimes c_\ell,
\end{equation}
where $\otimes$ denotes the outer product and $a_\ell, b_\ell,$ and $c_\ell$ are the columns of $A, B$, and $C$, respectively. $A$, $B$, and $C$ will be referred to as the NNCPD factors. A nonnegative approximation with fixed $r$ is obtained by approximately minimizing the reconstruction error between $X$ and the NNCPD reconstruction $\hat X =\sum \limits_{  \ell= 1}^{r} a_\ell \otimes b_\ell \otimes c_\ell$ among all the nonnegative vectors. When the reconstruction error vanishes we say that an exact rank-$r$ NNCPD is obtained. The nonnegative rank, denoted as $\rank_{+}(X)$, is the minimum integer $r*$ so that there exists an exact rank-$r^*$ NNCPD of $X$. In what follows, unless otherwise stated, when we refer to rank of a tensor we are referring to nonnegative rank and unless indicated otherwise, $\|\cdot\|$ indicates the Frobenius norm.

\subsubsection{Existence and Uniqueness of Rank-$r$ NNCPD} \label{sec:theory}

In \cite{qi2018briefintro}, the authors explore rank-$r$ NNCPD of tensors
whose rank is unknown, raising the question of when rank-$r$ approximations exist and are unique. The existence question is answered by the following proposition, but the uniqueness question has not yet been completely settled.

\begin{proposition}[\cite{qi2018briefintro}]\label{t:uniqueapprox} Let $X$ be a generic nonnegative tensor with $r < \mathrm{rank}_+(X)$, and $Y = \argmin_{\mathrm{rank}_+(Y) \leq r} \|X - Y\|$. Then $Y$ is unique and has $\mathrm{rank}_+(Y) = r$.
\end{proposition}

What is not known is whether or not the resulting rank-$r$ approximation $Y$ itself has a unique rank-$r$ decomposition. We do not know of any framework for which this uniqueness can be guaranteed. A series of results gives only partial answers. Kruskal's theorem \cite{kruskal} provides a test for 3-tensors based on the $k$-rank of the factor matrices of an NNCPD of the tensor. Generic results on spaces of tensors \cite{qi2018briefintro, domanov2014uniqueness} give partial results for spaces with specific conditions on tensor dimensions. The strongest of these restricts the product of the tensor dimensions to no more than 15,000 which is too small for most realistic cases, and imposes additional restrictions on tensor dimensions. Another strong result proves uniqueness for NNCPD for tensors of ranks of 2 or 3 and dimensions of at least 3. In what follows, we require existence but not uniqueness.

We also comment here that the choice of $r$ will of course affect the quality and interpretability of the output topics. Like in other classical NMF and related methods, this is a common challenge in determining a choice of $r$ that both allows for adequate topic representation while not being so large as to start fitting e.g., noise. Approaches include statistical parameter tuning  \cite{ulfarsson2013tuning,ito2016rank} as well as using hierarchical topic modeling to allow for various levels of granularity \cite{song2013hierarchical,cichocki2007hierarchical,GHMNSWZ19}. Applying these techniques to our approach is interesting future work.

\section{Comparison of NNCPD and NMF-based Nonnegative Tensor Decompositions} \label{sec:synthetic}

In this section, we perform numerical experiments showcasing how one might interpret the factors of NNCPD for dynamic topic modeling.
Each of these experiments highlights different features of NNCPD for topic modeling. In all the experiments, NNCPD outperforms Direct NMF and Fixed-factor NMF, as NNCPD provides more comprehensive analysis of the topic evolution while the other  methods fail to detect key changes. 
In all of our experiments, we use the tensorlab package~\cite{TensorLab} with \MATLAB. Specifically,  for NNCPD we use the sdf\_nls nonlinear least squares algorithm with default parameters and no regularization~\cite{sorber2015structured}, and for NMF we use \MATLAB's default implementation of NMF~\cite{berry2007algorithms}.

\subsection{Synthetic Dataset Numerical Experiments}
\label{sec:syntheticdataset}
In our first experiment, we consider numerical experiments on synthetic datasets where topic evolution is simple, and   only one topic changes (Sect.~\ref{sec:simple}). In our second experiment, we consider more complex data with repeated topic emergence, topic fading, and shifting topics (Sect.~\ref{ss: sinusoidal data}). 

\subsubsection{Monotonic Dynamic Topic Modeling  Dataset Experiment}
\label{sec:simple}
 
We consider a toy example of a $10\times 20\times 30$ 
nonnegative tensor $T$. 
We corrupt the original rank-3 tensor $X$ with noise, i.e., $T = X + 10^{-3} \cdot |Z|$, where the entries of the tensor $Z$ are sampled from the standard normal distribution. 
In an application, the tensor could represent a dataset whose first mode of dimension 10 corresponds to time, second mode of dimension 20 corresponds to survey questions, and the third mode of dimension 30 corresponds to users who answer those questions across time.

\begin{figure}[h!] 
	\centering
	\includegraphics[height=1.25in]{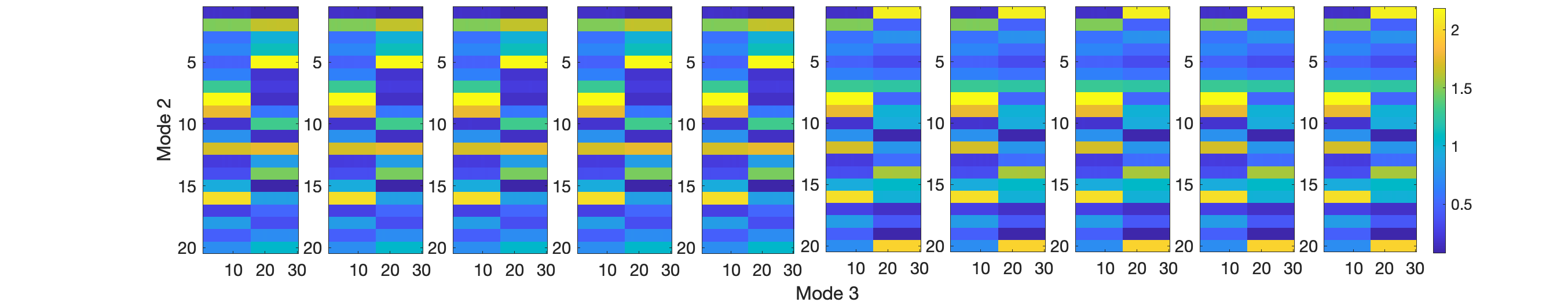} 
	\caption{Nonnegative ${10 \times 20 \times 30}$ tensor example constructed so that for all ten slices, the first $15$ columns are highly correlated.  
	For the first five slices, the second fifteen columns are also highly correlated, and the same for last five slices. Thus, for a given slice, the first 15 columns are correlated and the last 15 columns are correlated. 
	Across slices (time), a topic change has occurred between slice 5 and 6.}\label{fig1:T}
\end{figure}

In Figure \ref{fig1:T}, we show slices of the tensor across the first mode (time). This construction is a simple example of a situation when the first 15 users are very similar, and the second 15 users are very similar. For example, these two groups could correspond to healthy and sick patients. The healthy group answers all 20 questions quite similarly across time, which is evident by the correlation between the first 15 columns of each slice. The sick group also always answers the questions similarly, but those answers change drastically between time step 5 and 6 (perhaps after being given some medication, for example). That shift is captured by the change in the appearance of the last 15 columns between slices 5 and 6. Two questions we  wish to ask 
are (i) can we detect this topic change between time 5 and 6? (ii) can we identify that there are two core groups of patients? \begin{figure}[h!]
	\centering
	\includegraphics[height=2.5in]{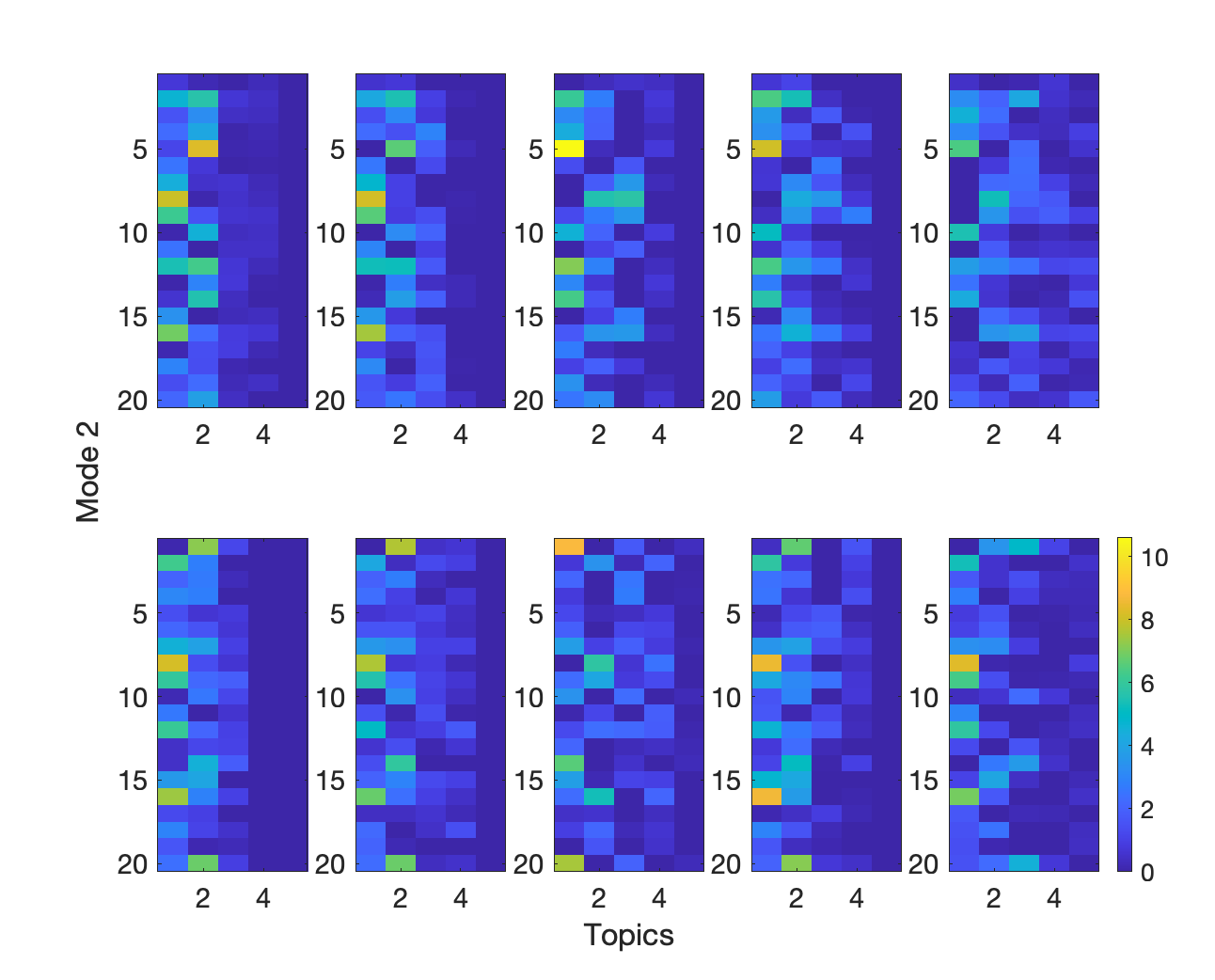} 
	\includegraphics[height=2.5in]{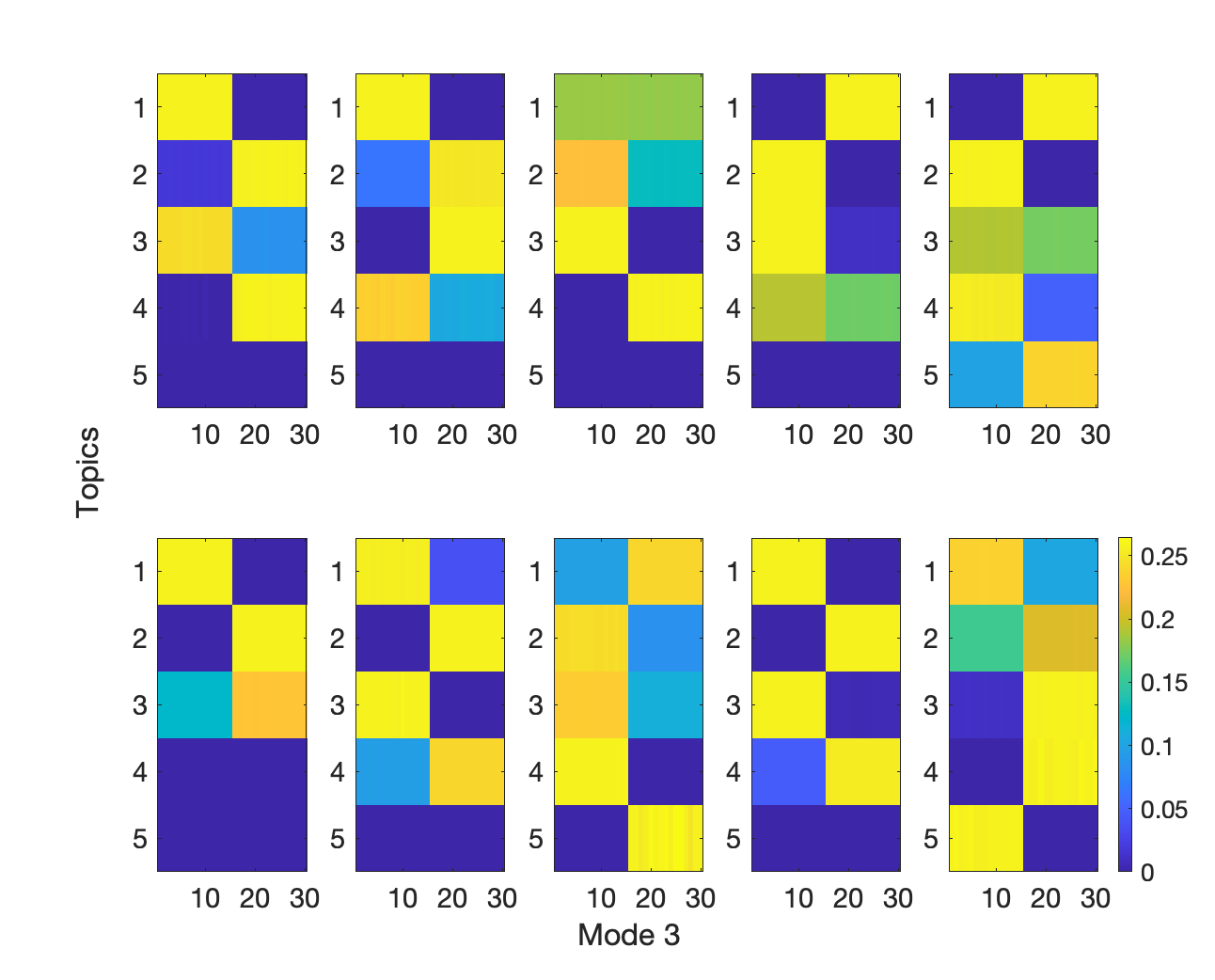} 
	\caption{Direct NMF performed 
		slice by slice 
		for tensor in Figure \ref{fig1:T}. Left: Direct NMF $A$ factors with $r=5$ topics. It appears that the first topic corresponds to the topic showcased by the first 15 columns of each tensor slice. Right: Direct NMF $S$ factors with $r=5$. Clearly, in each slice the first and last 15 columns are highly correlated. 
		Direct NMF reconstruction error is $\| T-\hat T\|_F = 30.1421$.
		}\label{fig1:W}
\end{figure}

\begin{figure}[h!]
	\centering
	\includegraphics[height=1.25in]{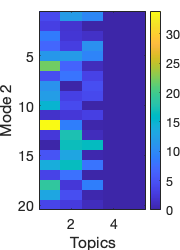} 
	\includegraphics[height=1.25in]{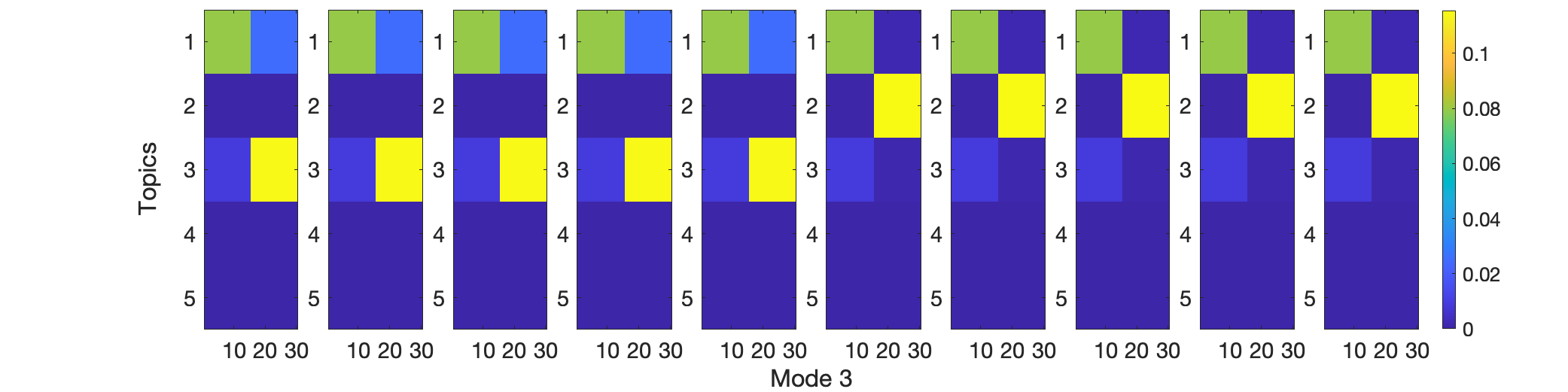} 
	\caption{Fixed NMF performed on matricized tensor for tensor in Figure \ref{fig1:T} for $r=5$. Left: Fixed NMF common $A$ factor for each slice. Right: Fixed NMF $S$ factors for each slice. Clearly, in each slice the first and last 15 columns are highly correlated. 
	The topic change is also evident between the fifth and sixth $S$ factors.  
	Fixed NMF reconstruction error is $\| T-\hat T\|_F = 2.6603$. 
	}\label{fig1:AS}
\end{figure}

\begin{figure}[h!]
	\centering
	\includegraphics[height=1.25in]{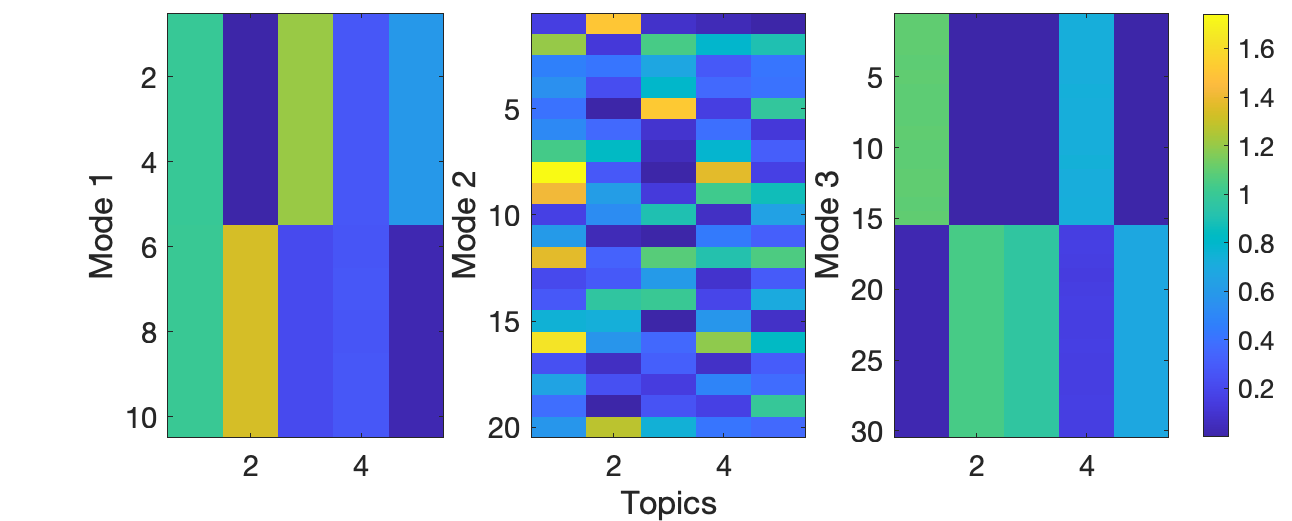} 
	\caption{NNCPD factors $A$, $B$, and $C$ (Left, Center, and Right, respectively) with rank 5 for the tensor from Figure \ref{fig1:T}. Notice that factor $A$ showcases topic evolution across time (slices). 
	The $B$ factor showcases the question-representation of topics (i.e., the answer patterns that are associated to each topic).
	The $C$ factor 
	showcases the user-representation of topics (i.e., the users associated with each topic).
	NNCPD reconstruction error is $\| T-\hat T\|_F = 0.045761$.
	}\label{fig1:ABC}
\end{figure}

To answer the above questions, we perform Direct NMF and Fixed NMF  slice by slice, slicing across the first mode  (note that we thus do not use any temporal correlation) to compute the  $A$ and $S$ factors. We also compute an NNCPD for the entire tensor, to obtain NNCPD factor matrices $A$, $B$, and $C$. Our results for this first  example are shown in Figures \ref{fig1:W}-\ref{fig1:ABC}.
The associated reconstruction errors are presented in the captions of the figures.
The reconstruction error in this section is defined as $\| T-\hat T\|_F$, where $\hat T$ is the reconstruction of the noisy tensor $T$ using the associated method.
This notion of reconstruction error is adapted in applications where the underlying true tensor is unknown.
As there is random noise added to the tensor, these errors could vary trial to trial.   
In Sect.~\ref{sec: robustness}, we explore the variance in the reconstruction error and its relationship to the size of the noise, where we define the reconstruction error as $\| X-\hat T\|_F$, and $\hat T$ is the reconstruction of the noisy tensor $T$ using the associated method.
This notion of reconstruction error will help us in understanding NNCPD as a denoising method, as we do not wish to fit the noise $N$.

In these experiments and those following, the reconstruction errors for Direct NMF, Fixed NMF, and NNCPD vary and often that of Direct NMF is greater than that of Fixed NMF, and both greater than that of NNCPD.  This may be surprising given the fact that Fixed NMFs are a subset of Direct NMFs, and NNCPD are a subset of both Direct NMFs and Fixed NMFs (using a matricization transformation).  Thus, we have the reverse ordering between the reconstruction errors of the global optima of each decomposition from what is seen experimentally.  However, we note that the objective functions defining each decomposition are nonconvex and the decompositions produced are therefore highly dependent on the method used for training and its initialization.  Furthermore, the decompositions produced in subsequent trials will vary in reconstruction error as they find different local minima.  Thus, this ordering of reconstruction errors need not hold for the local minima decompositions produced.  This highlights the need to measure the quality of such decompositions  by more than solely reconstruction error and motivates both our investigation of their interpretability via the ``eyeball" metric and our investigation of the robustness of each decomposition to noise in Section \ref{sec: robustness}.


Figure~\ref{fig1:ABC} displays the three factor matrices obtained by performing NNCPD with rank 5 on the 3-mode tensor in Figure~\ref{fig1:T}. 
The matrix $B\in \mathbb{R}^{20 \times 5}_{\ge 0}$ showcases the question-representation of topics, i.e., the answer patterns that are associated to each topic.
We refer to the columns of $B$, or the answer patterns, as topics.
For example in Figure~\ref{fig1:T}, the first fifteen users have very similar answer patterns throughout time.
We observe in the $B$ matrix of Figure~\ref{fig1:ABC}, that the first column captures this pattern, and identifies this pattern as one of the topics.
Further, matrix $C \in \mathbb{R}^{30 \times 5}_{\ge 0}$ showcases the user-representation of topics, i.e., the users associated with each topic. 
We see that the first column of the $C$ matrix in Figure~\ref{fig1:ABC}, correctly associates the first topic with only the first fifteen users.
Lastly, matrix $A \in \mathbb{R}^{10 \times 5}_{\ge 0}$ showcases the topic evolution through time, i.e., the presence, absence, and evolution of the topics through time.
For the example in Figure~\ref{fig1:T}, the first topic associated with the first fifteen users is persistent throughout all 10 time slices of the tensor $T$.
We see that the first column of the $A$ matrix in Figure~\ref{fig1:ABC} indicates that the first topic persists through time.

Similar analysis can be done on the remaining topics identified in the columns of the $B$ matrix.
We see that the second and third columns of matrix $C$ indicate that the topics captured in the second and third columns of $B$ are primarily associated with the second fifteen users.
Further, the second column of the $A$ matrix indicates that the second topic captured in $B$ is absent in the first five time slices, but appears at the sixth time slice and persist through time. 
In contrast, the third column of the $A$ matrix indicates that the third topic captured in $B$ is present in the first five time slices, but disappears for the last five time slices.
Indeed, these two topics seem to represent the remaining two answer patterns associated with the second fifteen users showcased in the tensor of Figure~\ref{fig1:T}.

Our first experiment illustrates how the factor matrices of NNCPD capture the topic representation along each mode.
The tensor in Figure~\ref{fig1:T} seems to showcase three significant answer patterns which are largely captured in applying NNCPD to the tensor.
In this experiment, we chose a rank of five in all of the decompositions, which is an overestimate of our guess of the number of significant topics in the data. The number of topics is often unknown, and thus estimated.
In Sect.~\ref{sec: robustness}, we study the performance of the three decompositions (Direct NMF, Fixed NMF, and NNCPD) in terms of reconstruction error $\|X - \hat T \|_F$ as we vary the input rank, and the form and magnitude of noise.
\begin{figure}[h!]
	\centering
	\includegraphics[height=1.25in]{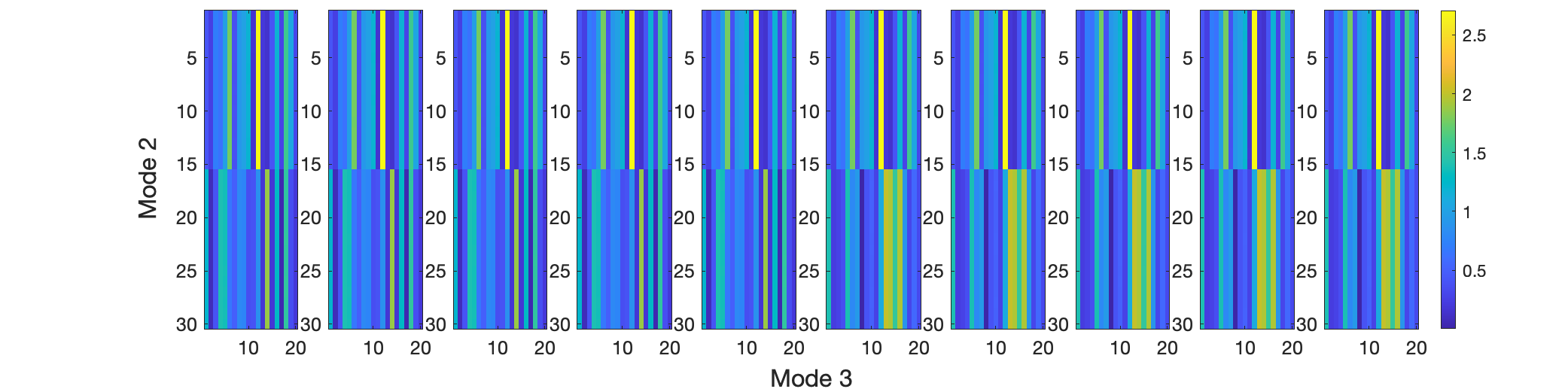}
	\caption{Nonnegative $ {10 \times 30 \times 20}$ tensor example constructed so that for all ten slices, the first $15$ rows are highly correlated. For the first five slices, the second fifteen rows are also highly correlated, and the same for the last five slices. Thus, for a given slice, the first 15 rows are correlated and the last 15 rows are correlated. Across slices (time), a topic change has occurred between slice 5 and 6.}\label{fig2:T}
\end{figure}

\begin{figure}[h!]
	\centering
	\includegraphics[height=2.5in]{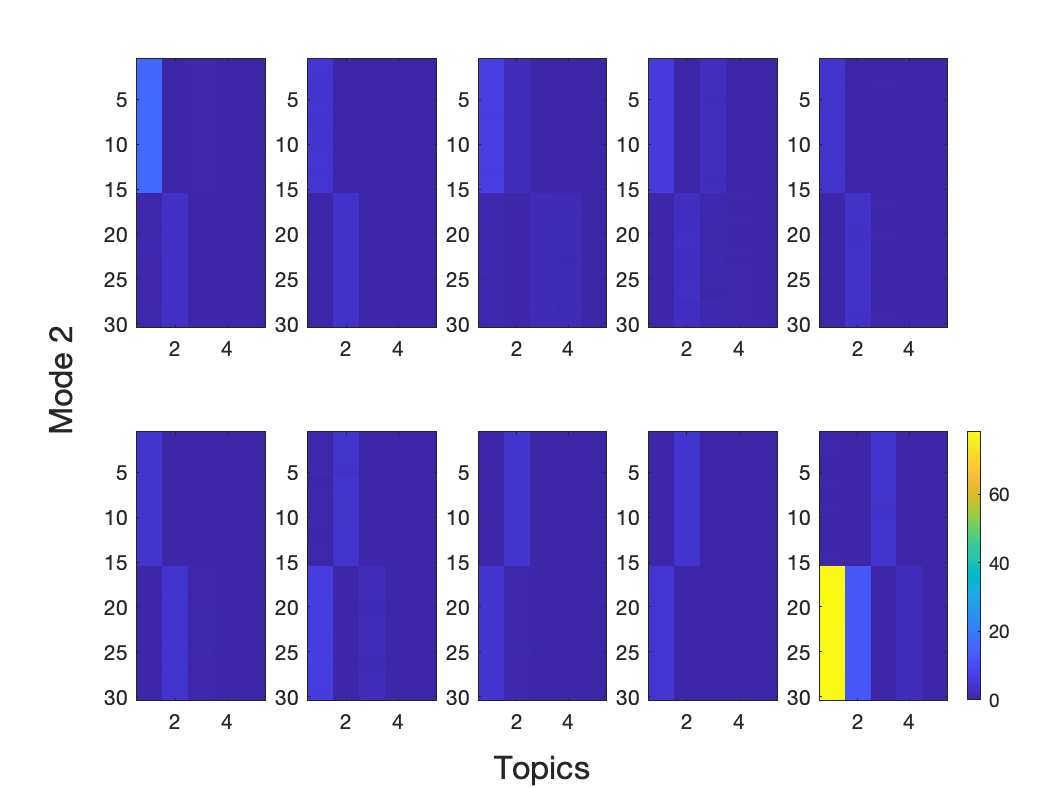} 
	\includegraphics[height=2.5in]{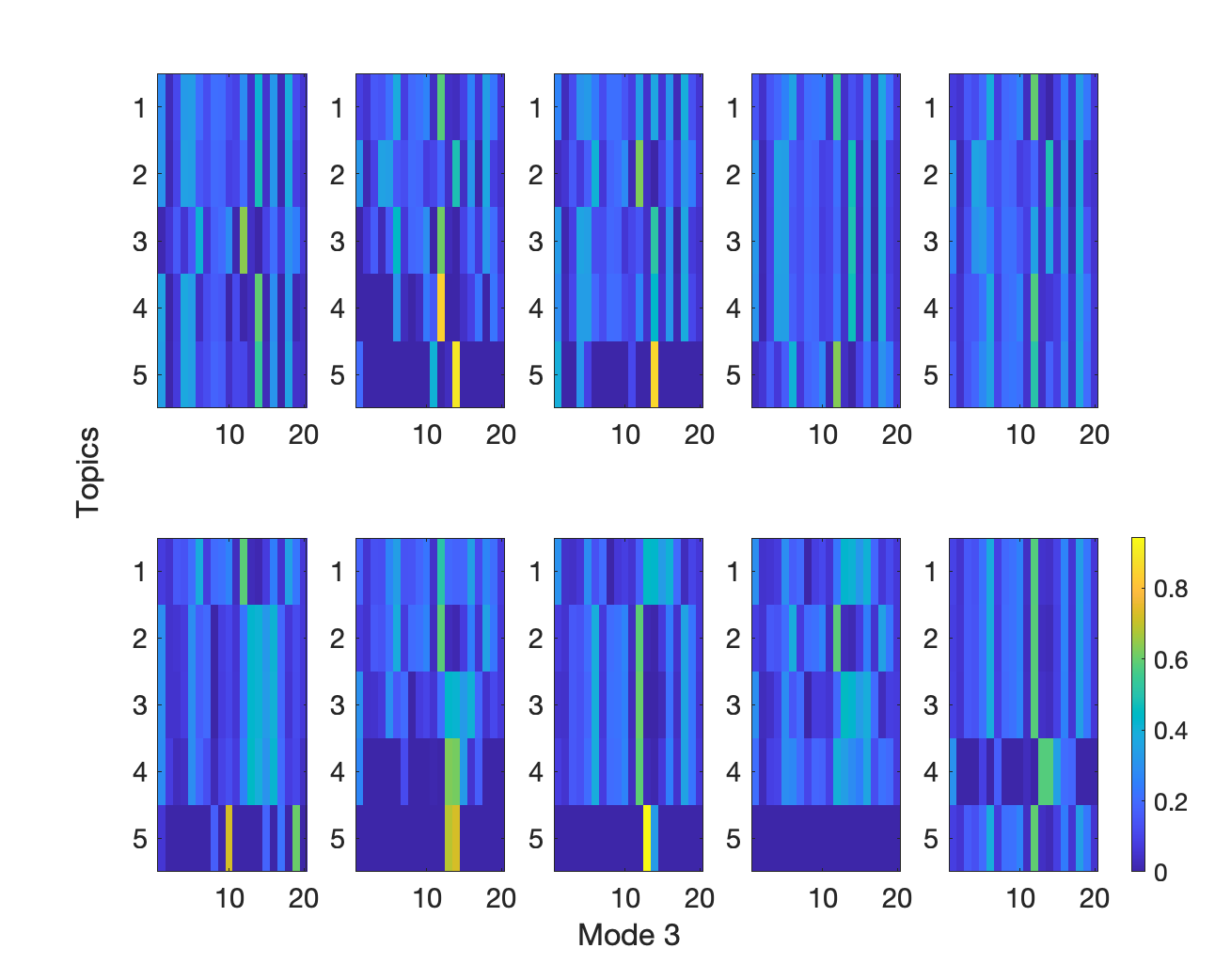} 
	\caption{Direct NMF performed 
		slice by slice 
		for tensor in Figure \ref{fig2:T}. Left: Direct NMF $A$ factors with $r=5$ topics. Right: Direct NMF $S$ factors with $r=5$. Clearly, in each slice the first and last 15 rows are highly correlated. 
		Direct NMF reconstruction error is $\| T-\hat T\|_F = 1031.9668$.
		}\label{fig2:W}
\end{figure}

\begin{figure}[h!]
	\centering
	\includegraphics[height=1.25in]{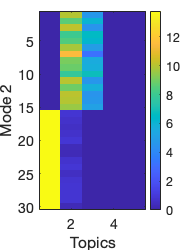}  
	\includegraphics[height=1.25in]{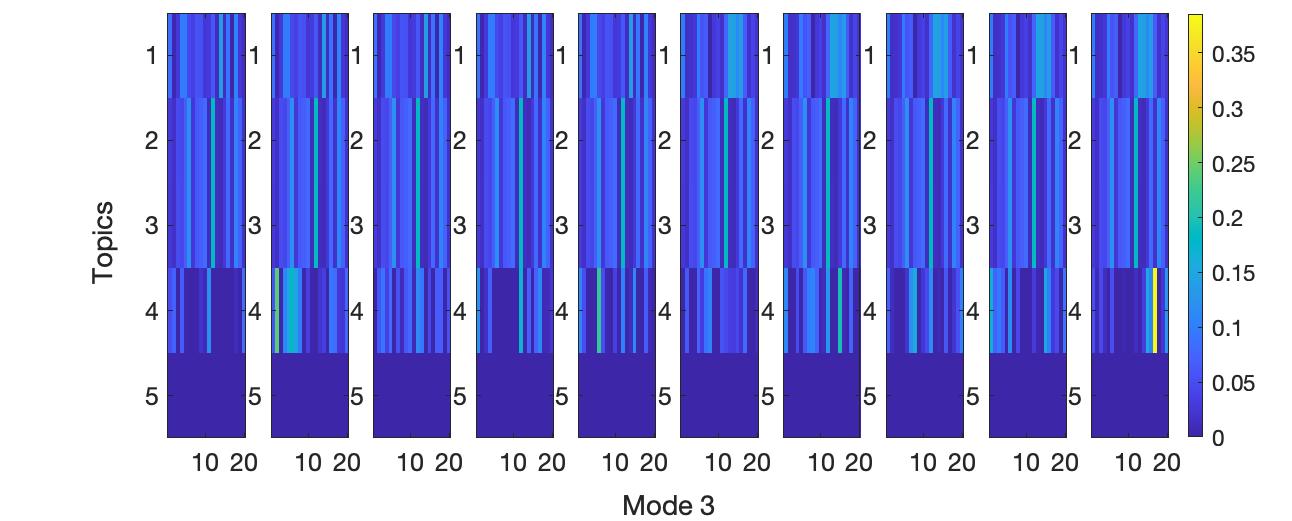} 
	\caption{Fixed NMF performed on matricized tensor for tensor in Figure \ref{fig2:T} for $r=5$. Left: Fixed NMF common $A$ factor for each slice. Right: Fixed NMF $S$ factors for each slice. The $A$ factor showcases the correlations between rows of the slices, but it is now harder to detect a topic change has occurred after time slice 5. 
	Fixed NMF reconstruction error is $\| T-\hat T\|_F = 2.1398$.
	}\label{fig2:AS}
\end{figure}

\begin{figure}[h!]
	\centering
	\includegraphics[height=1.25in]{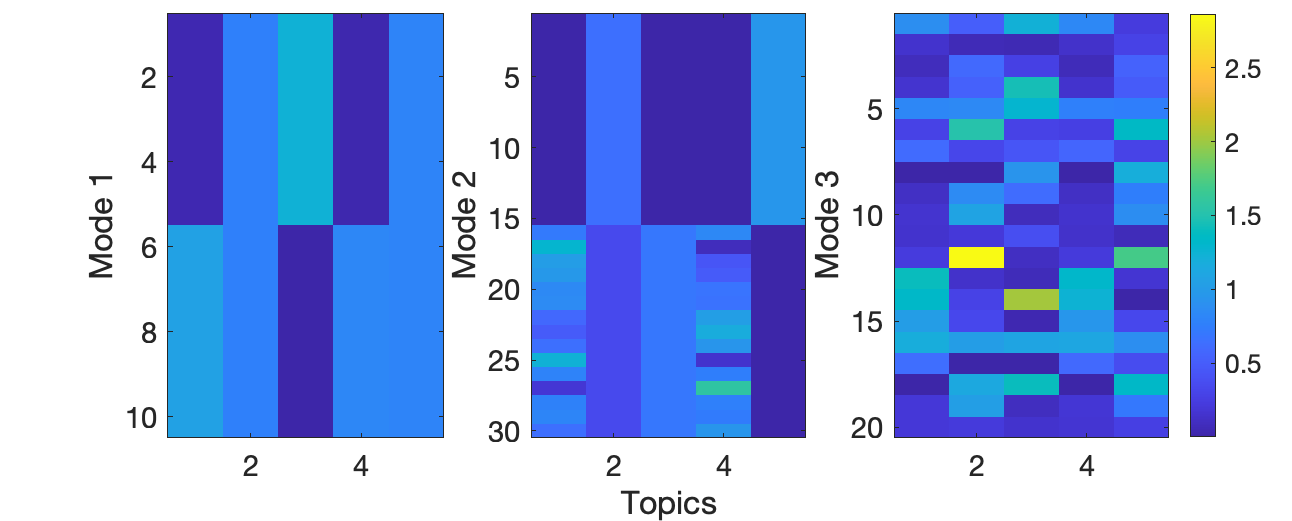}
	\caption{NNCPD factors $A$, $B$, and $C$ (Left, Center, and Right, respectively) with rank 5 for tensor from Figure \ref{fig2:T}. Notice that $A$ factor showcases topic evolution across time (slices). 
	The $B$ factor showcases the row-representation of topics i.e., the rows associated with each topic.
	The $C$ factor showcases the column-representation of topics i.e., the column patterns associated with each topic.
	NNCPD reconstruction error is $\| T-\hat T\|_F = 0.046283$.
	}\label{fig2:ABC}
\end{figure}

For the first example, it seems that both the Fixed NMF and our NNCPD method are able to highlight the topic shift. 
In contrast, when the data is transposed as a tensor with dimensions ${10 \times 30 \times 20}$ so that there is more variation in the third rather than second mode, this shift becomes harder to detect using the Fixed NMF approach. 
That is, as in Figure \ref{fig2:T}, we consider a ${10 \times 30 \times 20}$ tensor $T$. 
We corrupt the original rank-3 tensor $X$ with noise, i.e., $T = X + 10^{-3} \cdot |Z|$, where the entries of the tensor $Z$ are sampled from the standard normal distribution. 
The tensor is constructed so that for all ten slices, the first $15$ rows are highly correlated. For the first five slices, the second fifteen rows are also highly correlated, and the same for the last five slices. Thus, for a given slice, the first 15 rows are correlated and the last 15 rows are correlated. Across slices (time), a topic change has occurred between slice 5 and 6. 
Figures \ref{fig2:W}-\ref{fig2:ABC} show for the results of the Direct NMF, Fixed NMF and NNCPD performed on the transposed tensor.

Figure~\ref{fig2:ABC} displays the three factor matrices obtained by performing NNCPD with rank 5 on the 3-mode tensor in Figure~\ref{fig2:T}. 
The matrix $C \in \mathbb{R}^{20 \times 5}_{\ge 0}$ showcases the column-representation of topics i.e., the column patterns associated with each topic. 
We refer to the columns of $C$ as topics.
For example in Figure~\ref{fig2:T}, the first fifteen rows have very similar column patterns throughout time.
We observe in the $C$ matrix of Figure~\ref{fig2:ABC} that the second column captures this pattern, and identifies it as one of the topics.
Further, the matrix $B\in \mathbb{R}^{30 \times 5}_{\ge 0}$ showcases the row-representation of topics, i.e., the rows associated with each topic.
We see that the second column of the $B$ matrix in Figure~\ref{fig2:ABC} correctly associates the second topic primarily with the first fifteen rows.
Finally, the matrix $A \in \mathbb{R}^{10 \times 5}_{\ge 0}$ showcases the topic evolution through time.
For the example in Figure~\ref{fig2:T}, the topic associated with the first fifteen rows is persistent throughout all ten time slices of the tensor $T$.
We see that the second column of the $A$ matrix in Figure~\ref{fig2:ABC} indicates that the second topic does indeed persist through time.

Similar analysis can be done on the remaining topics identified in the columns of the $C$ matrix.
We see that the third column of matrix $B$, indicates that the topic captured in the third column of $C$ is primarily associated with the second fifteen rows.
Further, the third column of the $A$ matrix indicates that the third topic captured in $C$ is present in the first five time slices, but disappears in the last five time slices.
Indeed, this topic represents the column patterns associated with the second fifteen rows showcased in the first 5 time slices of the tensor in Figure~\ref{fig2:T}.

Furthermore, we observe that the first and fourth columns $C$ are very similar.
The matrix $B$ indicates that the topics captured in the first and fourth columns of $C$ are primarily associated with the second fifteen rows.
Further, the first and fourth columns of the $A$ matrix indicate that the corresponding topics are not present in the first five time slices, but are present for the last five time slices.
These topics represent the column patterns associated with the second fifteen rows showcased in the last five time slices of the tensor in Figure~\ref{fig2:T}.

Note that when the data is transposed in the second experiment the topic shift becomes harder to detect using Fixed NMF.
\textit{\bfseries Therefore, without knowing a priori the structure of the dynamic component (e.g., what mode it lies in), we have no reason to believe these prior approaches will be able to detect such an event. NNCPD on the other hand, clearly and easily highlights such changes along any mode.} We emphasize that this is only a \textit{visual} observation, but that future work could develop an analysis of these patterns to automate the identification of topic changes, as in \cite{chen2015modeling}. One could, as a simple approach, compute the temporal gradient of the output factors and identify the largest spikes as corresponding to topic shifts.

\subsubsection{Complex Dynamic Topic Modeling Dataset Experiment} \label{ss: sinusoidal data}

In this section we consider synthetic tensor datasets with more complicated topic evolution than those in Sect.~\ref{sec:simple}, such as emergence, fading, and shifting.

\subsubsection*{Topic Emergence and Fading Experiment}
\begin{figure} [h!]
	\centering
	\includegraphics[height=2.5in]{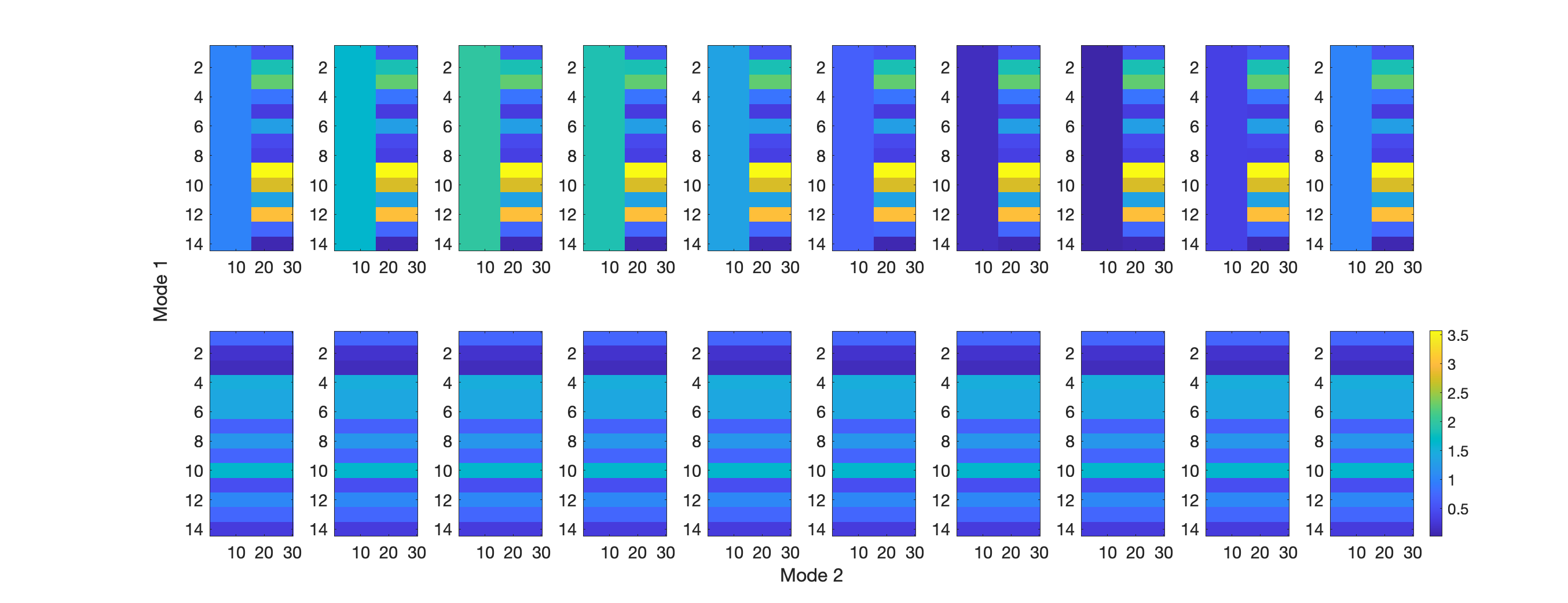} 
	\caption{Tensor example $X \in \R_{\geq 0}^{14\times 30 \times 20}$ is constructed so that in the first ten slices of the tensor, the first $15$ columns are highly correlated and the second $15$ columns are highly correlated. Further, in the second ten slices, all $30$ columns are highly correlated. Further, across slices, sudden topic changes have occurred between slices 10 and 11.
	}\label{fig1:T_exp1}
\end{figure}

We now consider a synthetic tensor data that models situations where a topic emerges, fades, then re-emerges. We construct a nonnegative tensor $X \in \Re_{\geq 0}^{14 \times 30 \times 20}$ as follows,
\[ X_{ijk} = \begin{cases} 
\sin \left((k-1)(2\pi/9) \right) + 1 &\mbox{for } i = 1, \cdots,14, \quad j = 1, \cdots 15, \mbox{ and } k = 1, \cdots 10 \\
|z_i| &\mbox{for } i = 1, \cdots,14 , \quad j = 15, \cdots 30, \mbox{ and } k = 1, \cdots 10 \\
|w_i| &\mbox{for } i = 1, \cdots,14 , \quad j = 1, \cdots 30, \mbox{ and } k = 10, \cdots 20,
\end{cases}
\]
where $z_i$ and $w_i$, defined for each $i$, are drawn from the standard normal distribution. See   Figure~\ref{fig1:T_exp1} where we show slices across the third mode, the temporal dimension. 

\begin{figure} [h!]
	\centering
	\includegraphics[height=2.1in]{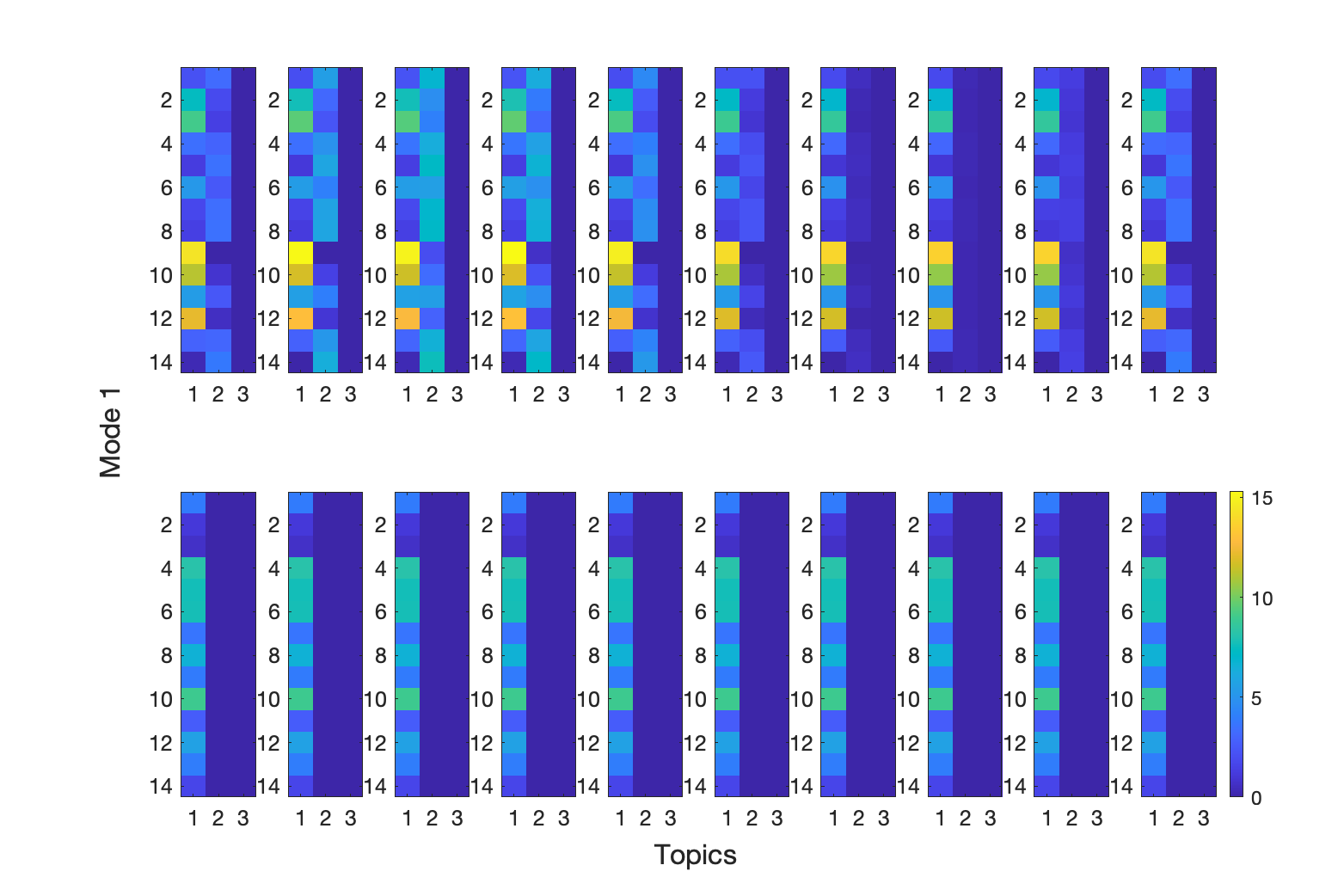} 
	\includegraphics[height=2.1in]{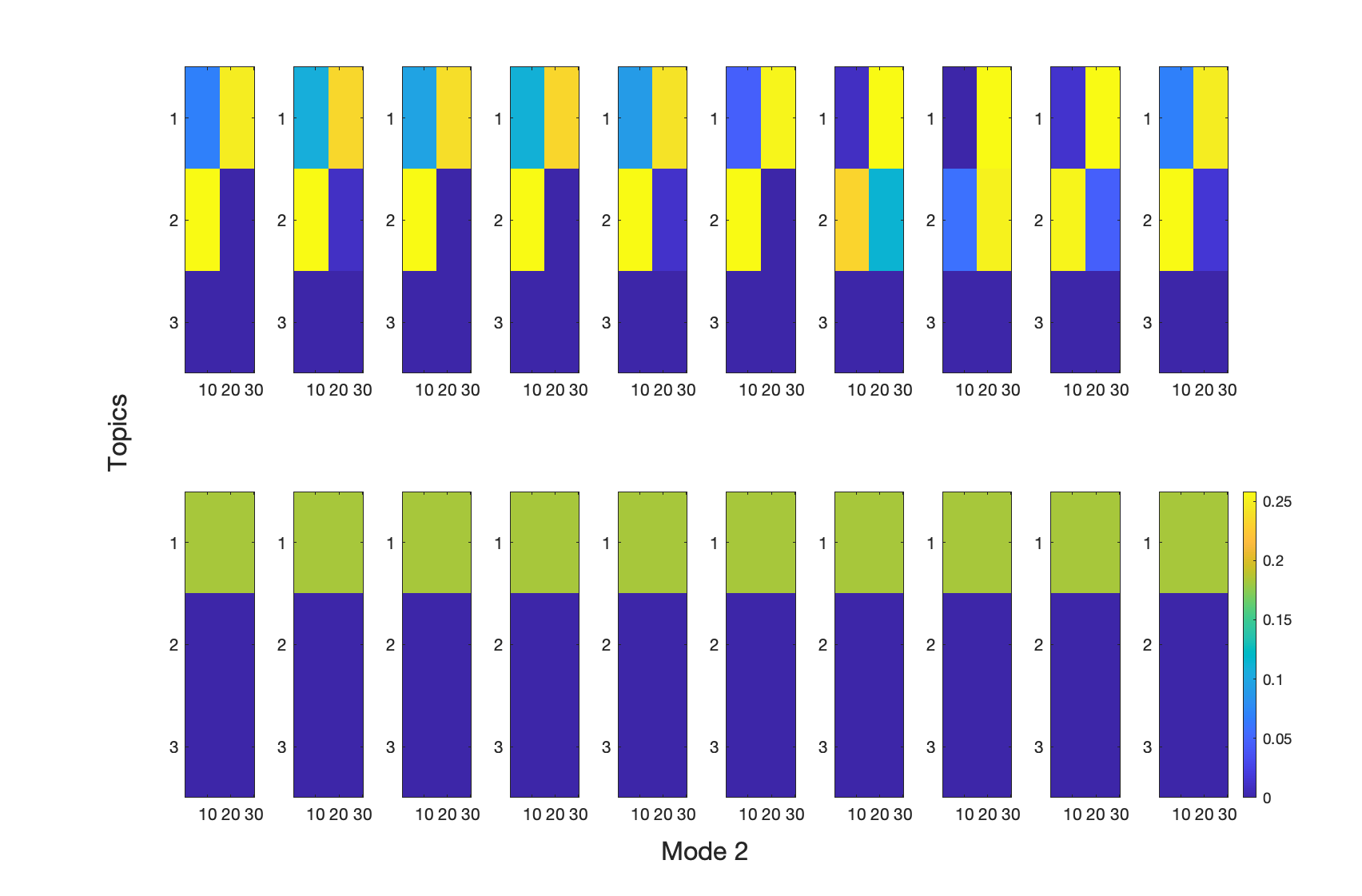} 
	\caption{Direct NMF performed slice by slice for tensor in Figure \ref{fig1:T_exp1} with $r=3$ topics. Left: Direct NMF $A$ factors. It appears that the first topic (in the $A$ factor) can be seen in the second 15 columns of the first ten slices of the tensor $T$. Right: Direct NMF $S$ factors aligned with the $A$ factors. In each of the first ten slices, the first and last 15 columns are highly correlated. 
	Direct NMF reconstruction error, defined in Equation~\ref{eq:rec_error_fixed_nmf}, is 0.021549.}\label{fig1:W_exp1}
\end{figure}

\begin{figure} [h!]
	\centering
	\includegraphics[height=2.5in]{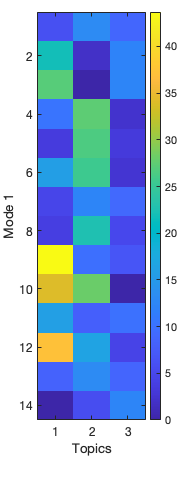} 
	\includegraphics[height=2.5in]{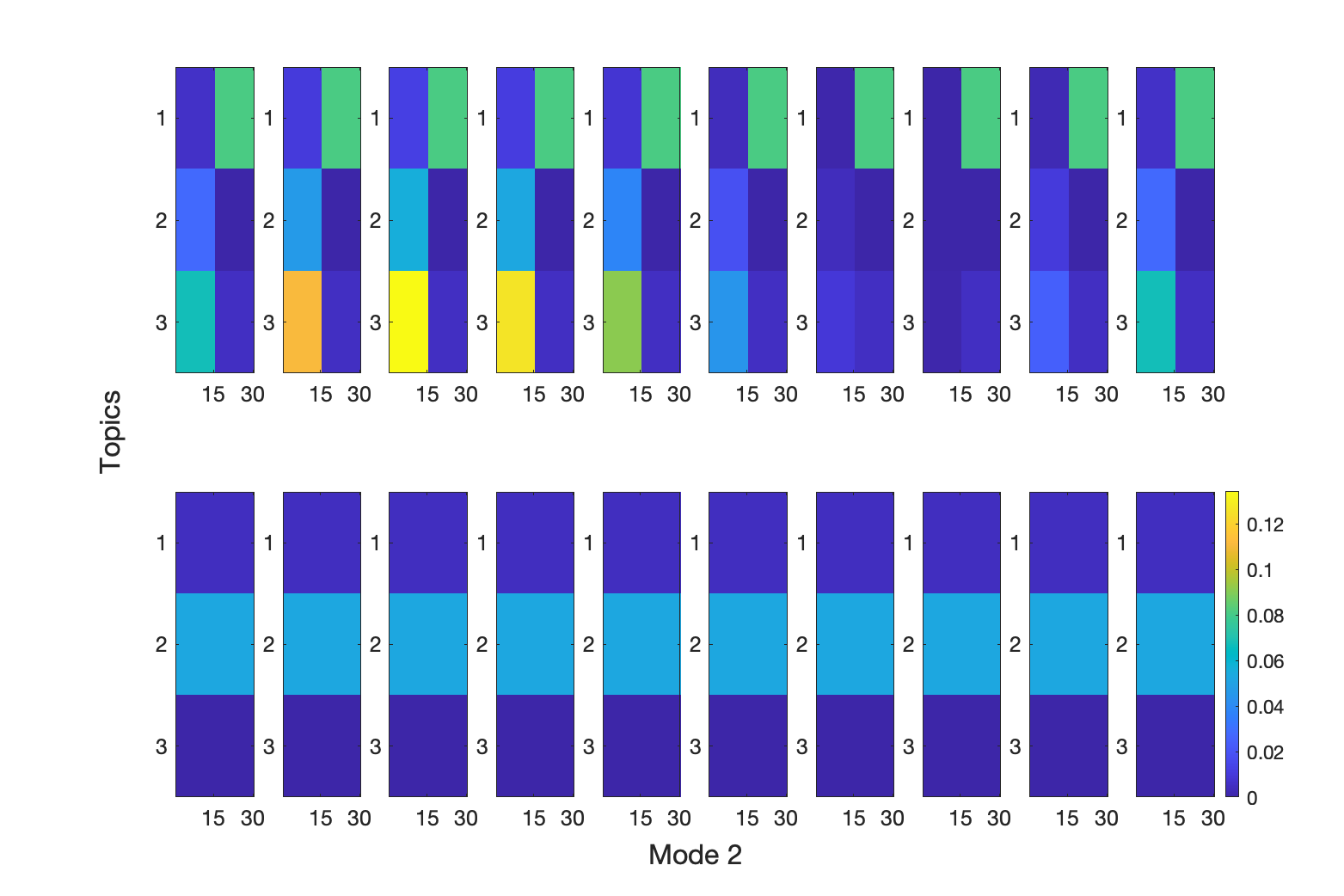} 
	\caption{Fixed NMF performed on matricized tensor for tensor in Figure~\ref{fig1:T_exp1} for $r=3$. Left: Fixed NMF common $A$ factor for each slice. Right: Fixed NMF  $S$ factors for each slice. In the right plot, for each of the first ten slices the first and last 15 columns are highly correlated and the topic change is also evident between slices 10 and 11. 
	Fixed NMF reconstruction error is $\| X - \hat X \|_F = 0.065896$.}\label{fig1:AS_exp1}
\end{figure}

\begin{figure} [h!]
	\centering
	\includegraphics[height=1.25in]{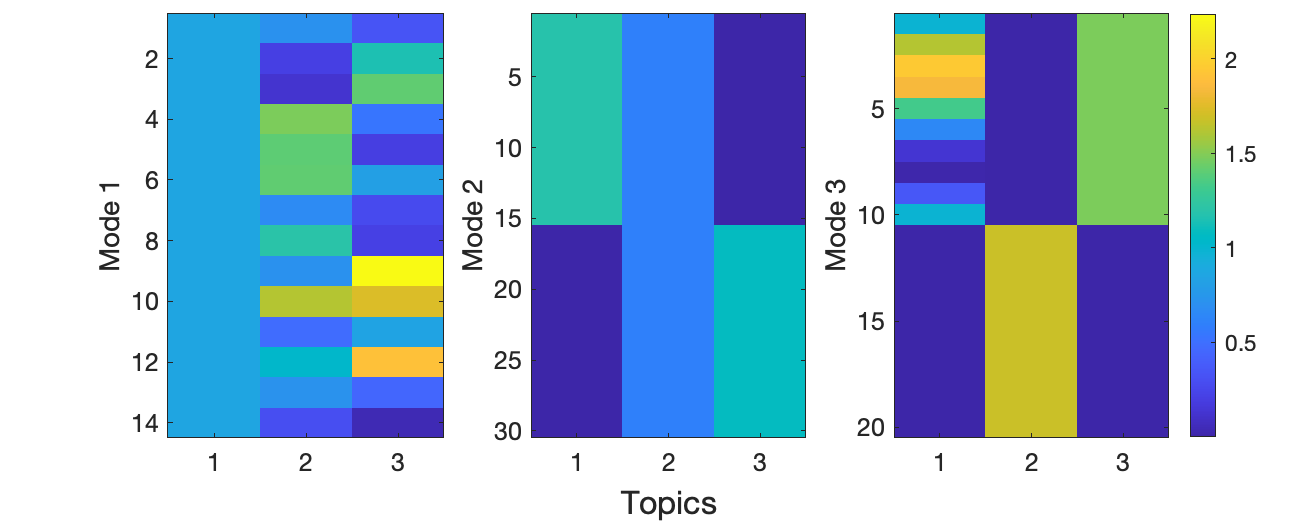} 
	\caption{NNCPD factors $A$, $B$, and $C$ (Left, Center and Right, respectively) with rank 3 for the tensor from Figure \ref{fig1:T_exp1}. The $A$ factor seems to indicate the topics; for instance, the third topic can be seen in the second 15 columns of the first ten slices of the tensor.  The $B$ factor showcases, among other things, that the third topic seems to correspond to the second 15 columns (in each of the first ten slices of the tensor using the factor $C$). Notice that factor $C$ showcases topic evolution across time (mode 3). Further, there is a clear event between slices 10 and 11. NNCPD reconstruction error is $\|X - \hat X \|_F = 0.0025301$.}\label{fig1:ABC_exp1}
\end{figure}

In applications, this dataset might be a situation where symptoms fade away then reappear, patterns are periodic, or topics become unpopular for awhile then trend again. We perform Direct NMF and Fixed NMF by slicing the tensor across mode 3, the temporal dimension, as displayed in Figures~\ref{fig1:W_exp1} and~\ref{fig1:AS_exp1} respectively.
We then compute a NNCPD for the entire tensor, to obtain NNCPD factors $A$, $B$, and $C$ displayed in Figure~\ref{fig1:ABC_exp1}.
The reconstruction errors $\|X - \hat X \|_F$ are presented in the captions of the figures, where $\hat X$ is the reconstruction of $X$ using the associated method.

Figure~\ref{fig1:ABC_exp1} displays the three factor matrices obtained by performing NNCPD with rank 3 on the 3-mode tensor in Figure~\ref{fig1:T_exp1}. 
The matrix $A \in \mathbb{R}^{14 \times 3}_{\ge 0}$ showcases the row-representation of topics i.e., the row patterns associated with each topic. 
We refer to the columns of $A$ as topics.
For example in Figure~\ref{fig1:T_exp1}, the second fifteen columns in the first ten time slices have the same row patterns.
We observe in the $A$ matrix of Figure~\ref{fig1:ABC_exp1} that the third column captures this pattern, and identifies it as one of the topics.
Further, the matrix $B\in \mathbb{R}^{30 \times 3}_{\ge 0}$ showcases the column-representation of topics i.e., the columns associated with each topic.
We see that the third column of the $B$ matrix in Figure~\ref{fig1:ABC_exp1} correctly associates the third topic with the second fifteen columns.
Lastly, the matrix $C \in \mathbb{R}^{20 \times 3}_{\ge 0}$ showcases the topic evolution through time.
For example in Figure~\ref{fig1:T_exp1}, the topic associated with the second fifteen columns is persistent throughout the first 10 time slices of the tensor $T$.
We see that the third column of the $C$ matrix in Figure~\ref{fig1:ABC_exp1} indicates that the third topic is persistent through the first 10 time slices.

Similarly, in Figure~\ref{fig1:T_exp1}, the first fifteen columns in the first ten time slices have the same row values; however, these values evolve through time according to the sine function, representing a uniform pattern that emerges, fades, then re-emerges.
We observe in the $A$ matrix of Figure~\ref{fig1:ABC_exp1} that the first column captures this uniform pattern, and identifies it as one of the topics.
Further, we see that the first column of the $B$ matrix in Figure~\ref{fig1:ABC_exp1} correctly associates the first topic with the first fifteen columns.
Lastly, the third column of the $C$ matrix in Figure~\ref{fig1:ABC_exp1} clearly captures the dynamics and evolution of the first topic throughout time.
Similar analysis can be performed for the second topic captured in the second column of the $B$ matrix in Figure~\ref{fig1:ABC_exp1}; it is associated with all of the columns of the last ten time slices of the tensor in Figure~\ref{fig1:T_exp1}.

Indeed, NNCPD provides superior visualization of topic evolution through time, as demonstrated in the factor matrices of the decomposition in Figure~\ref{fig1:ABC_exp1}, where the NNCPD $A$ factor shows the topic representation for each 
row,  the $B$ factor displays topic representation for each 
column,  and the $C$ factor displays the evolution of the topics through time. \textbf{\emph{NNCPD alone succeeds in detecting all the events in the topic evolution.  The temporal factor $C$ illustrates which topics persist throughout time and which are emerging and fading. }}

\subsubsection*{Topic Shift Experiment}

We next consider a data tensor that can model a situation where an event happens that shifts the topics discussed by users, such as an election that shifts the topics discussed by political parties before and after.
\begin{figure}[h!]
	\centering
	\includegraphics[height=2.5in]{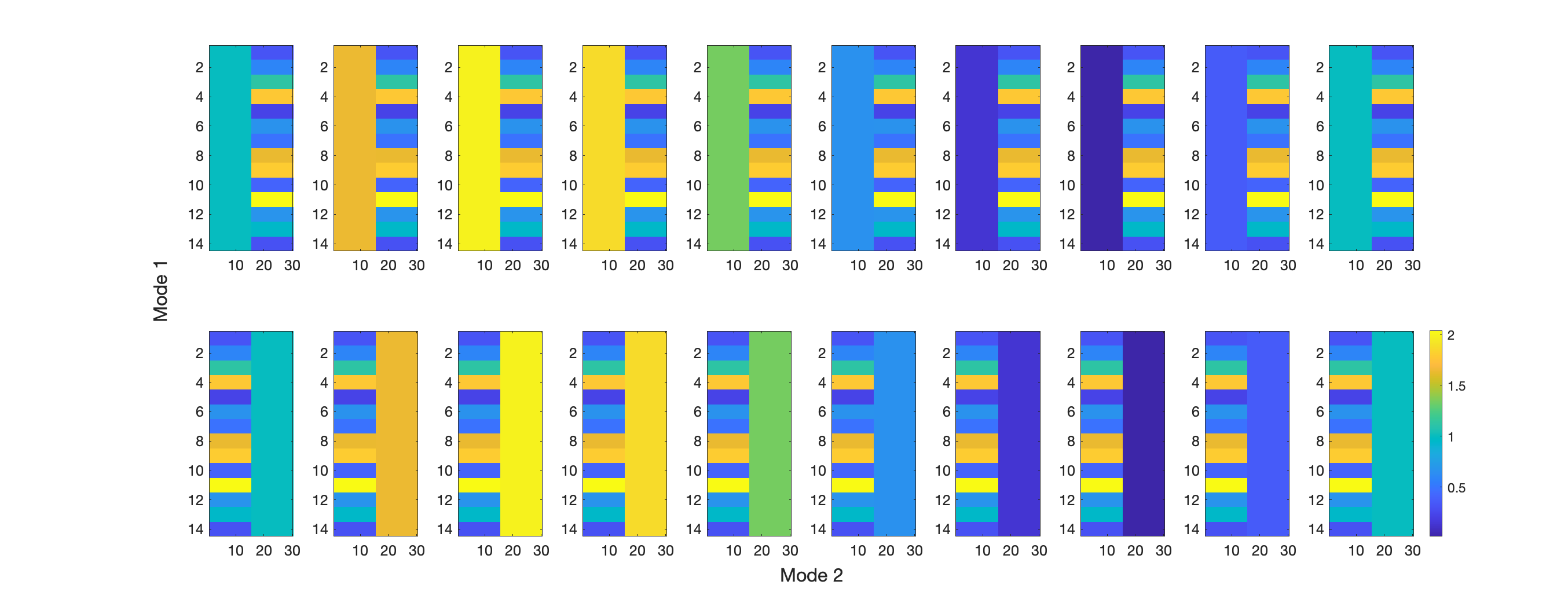}
	\caption{Tensor example $X \in \R_{\geq 0}^{14\times 30 \times 20}$ sliced along mode 3, the temporal dimension. In all slices, the first $15$ columns are highly correlated and the second $15$ columns are highly correlated. Across time, topic changes have occurred between slice 10 and 11.}\label{fig1:T_exp2}
\end{figure}
We construct a nonnegative data tensor $X \in \Re_{\geq 0}^{14 \times 30 \times 20}$ as follows,
\[ T_{ijk} = \begin{cases} 
\sin \left((k-1)(2\pi/9) \right) + 1& \mbox{for } i = 1, \cdots,14, \quad j = 1, \cdots 15, \mbox{ and } k = 1, \cdots 10 \\
|z_i| & \mbox{for } i = 1, \cdots,14 , \quad j = 15, \cdots 30, \mbox{ and } k = 1, \cdots 10 \\ 
\sin \left((k-1)(2\pi/9) \right) + 1 & \mbox{for } i = 1, \cdots,14, \quad j = 15, \cdots 30, \mbox{ and } k = 10, \cdots 20 \\
|z_i| & \mbox{for } i = 1, \cdots,14, \quad j = 1, \cdots 15, \mbox{ and } k = 10, \cdots 20
\end{cases}
\]
where $z_i$, defined for each $i$, is drawn from the standard normal distribution. See Figure~\ref{fig1:T_exp2} where we show slices across the third mode, the temporal dimension. 

\begin{figure}[h!]
	\centering
	\includegraphics[height=2.1in]{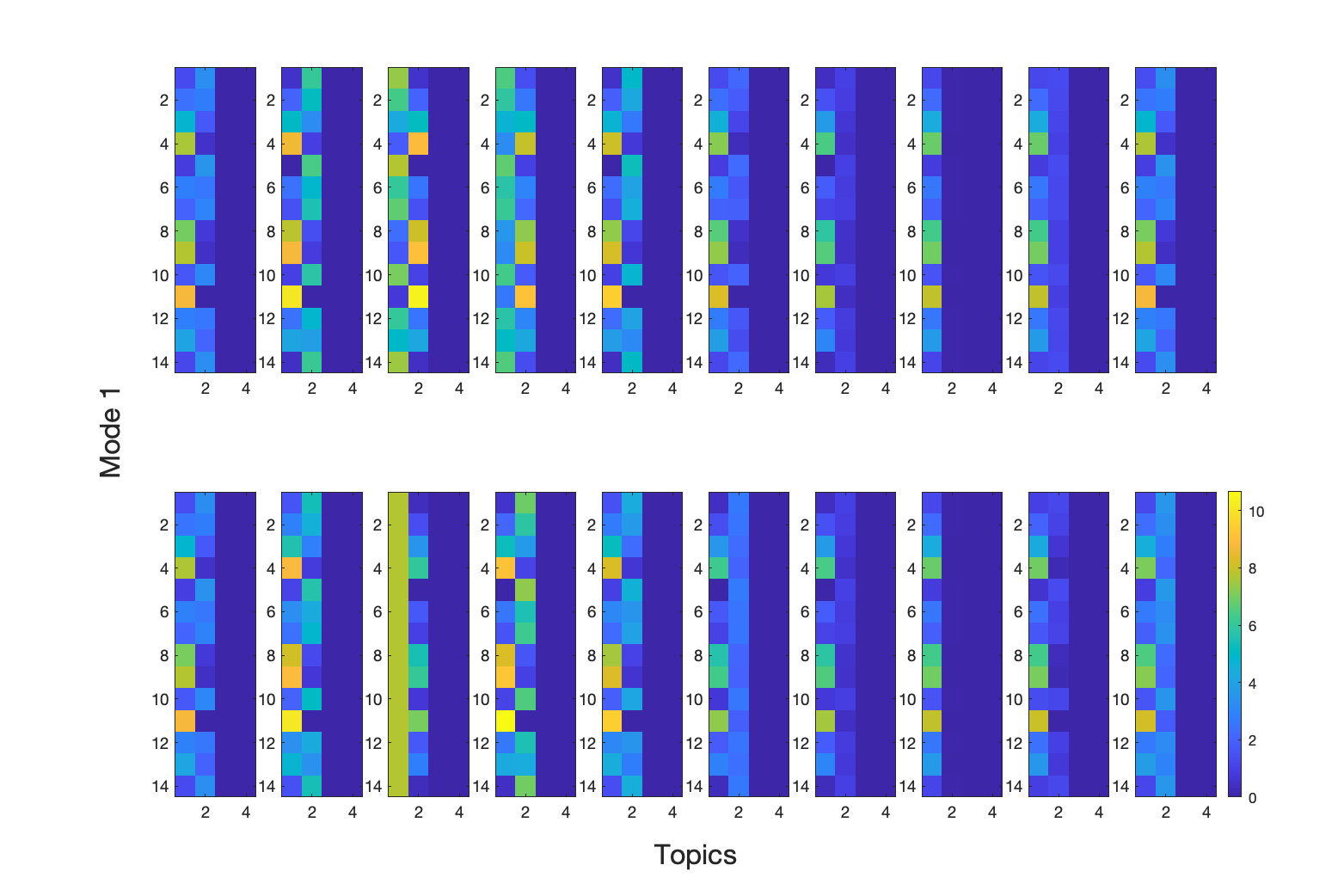}
	\includegraphics[height=2.1in]{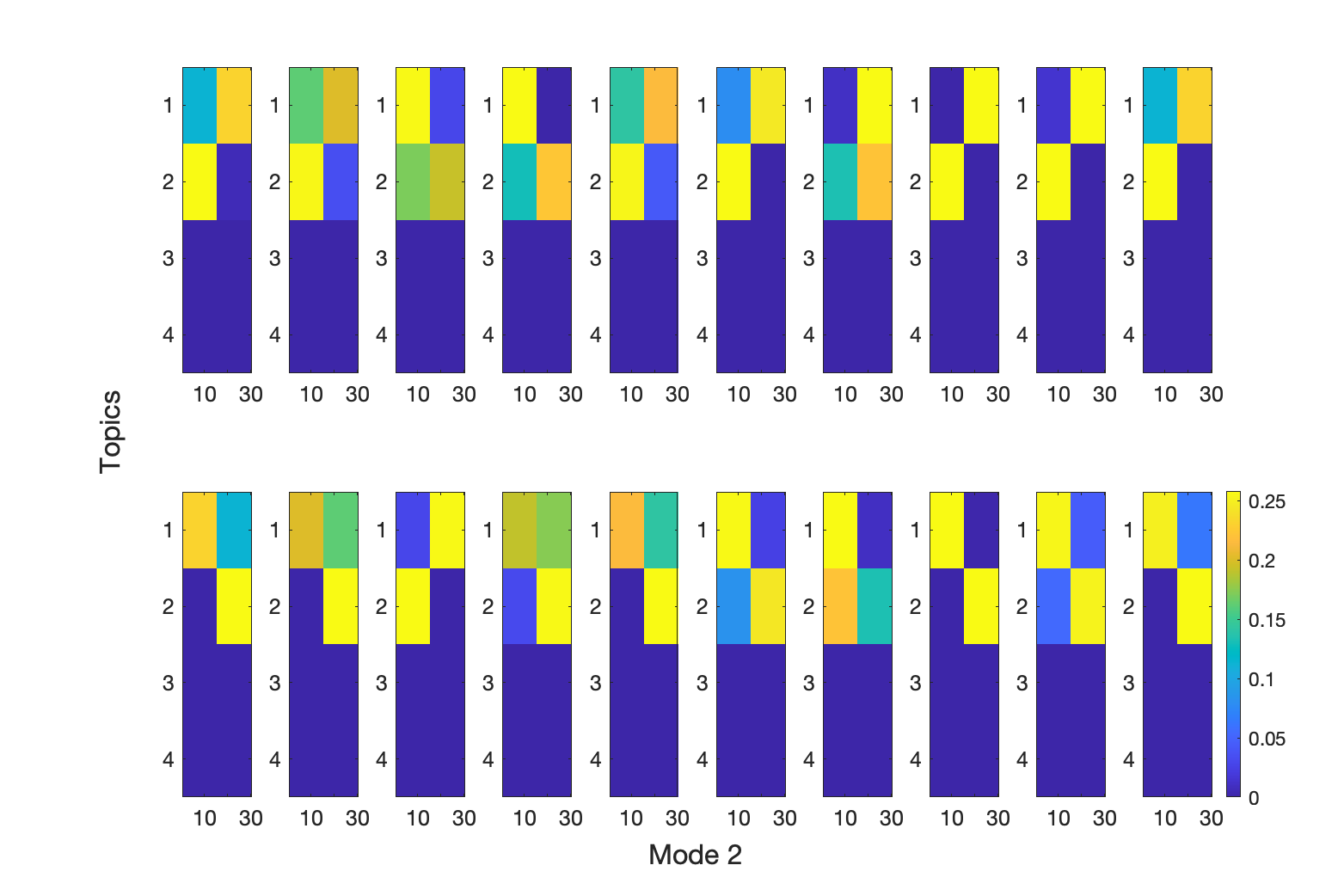}
	\caption{Direct NMF performed slice by slice for tensor in Figure \ref{fig1:T_exp2} with $r=4$ topics.  Left: Direct NMF $A$ factors. It appears that the first topic corresponds to the topic showcased by the second $15$ columns of each of the first $10$ time slices and consequently in the first $15$ columns of each of the second $10$ tensor slices. Right: Direct NMF $S$ factors aligned with the $A$ factors.  In all of the slices, the first and last 15 columns are clearly highly correlated. Direct NMF reconstruction error is $\|X - \hat X \|_F = 0.029248$.}\label{fig1:W_exp2}
\end{figure}

\begin{figure}[h!]
	\centering
	\includegraphics[height=2in]{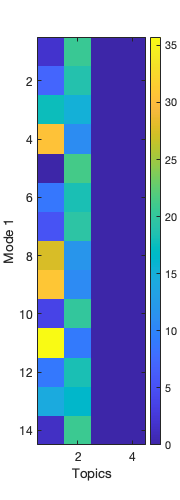}
	\includegraphics[height=2in]{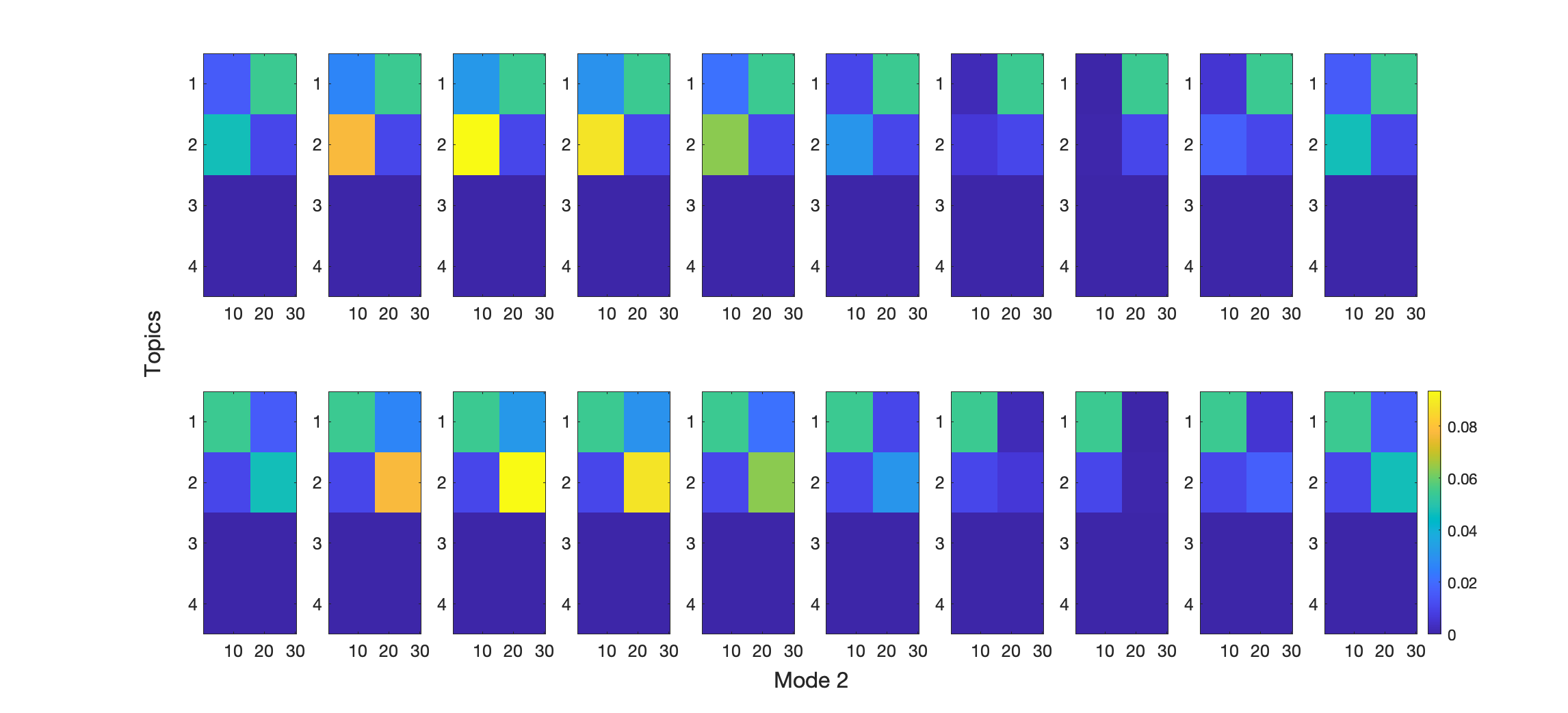}
	\caption{Fixed NMF performed on matricized tensor for tensor in Figure \ref{fig1:T_exp2} for $r=4$. Left: Fixed NMF common  $A$ factor for each slice. Right: Fixed NMF $S$ factors for each slice. Clearly, in all of the slices the first and last 15 columns are highly correlated. 
	Fixed NMF reconstruction error is $\|X - \hat X \|_F = 0.045349$.}\label{fig1:AS_exp2}
\end{figure}

\begin{figure}[h!]
	\centering
	\includegraphics[height=1.25in]{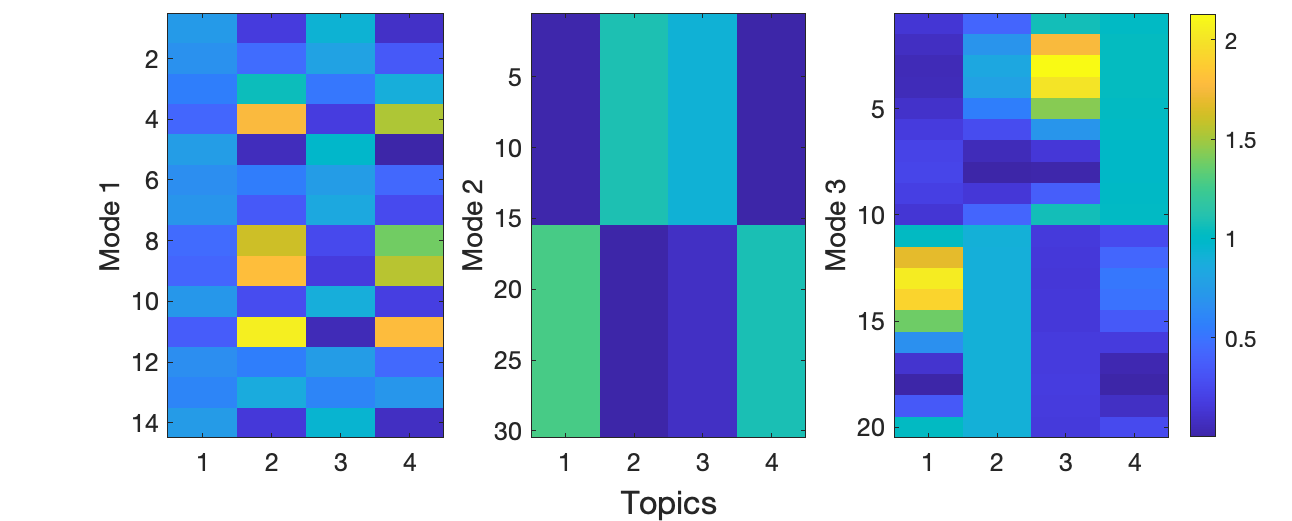}
	\caption{NNCPD factors $A$, $B$, and $C$ (Left, Center and Right, respectively) with rank equals 4 for the tensor from Figure \ref{fig1:T_exp2}.   NNCPD $A$ factor seems to indicate the topics; for instance, the second topic can be seen in the second 15 columns of the first ten slices of the tensor.    NNCPD $B$ factor showcases, among other things, that the second topic seems to correspond to the the first 15 columns in each of the first ten slices of the tensor. Notice that NNCPD  $C$ factor showcases topic evolution across the temporal dimension, mode 3. There is a clear event between slices 10 and 11. NNCPD reconstruction error is $\|X - \hat X \|_F = 0.00054029$.}\label{fig1:ABC_exp2}
\end{figure}

We perform Direct NMF and Fixed NMF by slicing the tensor along mode 3, the temporal dimension, as displayed in Figures~\ref{fig1:W_exp2} and~\ref{fig1:AS_exp2} respectively. 
We then compute a NNCPD for the entire tensor, to obtain NNCPD factors $A$, $B$, and $C$ displayed in Figure~\ref{fig1:ABC_exp2}. 
The reconstruction errors $\|X - \hat X \|_F$ are presented in the captions of the figures, where $\hat X$ is the reconstruction of $X$ using the associated method. 

We observe an additional strength of NNCPD for topic modeling. 
We know a priori that there are two topics (i.e., row patterns) in this tensor,  although the rank is $4$ (an example where the number of topics is different than the rank of the tensor).
We observe in Figure~\ref{fig1:ABC_exp2} how NNCPD can roughly detect and showcase both of these facts through its factor matrices. 
The NNCPD $A$ factor displays $4$ different topics, but the first and third are very similar, and the second and fourth are also very similar.  
Furthermore, NNCPD $B$ factor indicates which columns are associated with the topics, and the $C$ factor showcases the topics evolution through time.
For example, we observe in the $B$ and $A$ factor matrices that the first topic in matrix $C$ is associated with the last fifteen columns, and was not present in the first ten time slices, but evolved through time according to the sine function, representing a uniform pattern that emerges, fades, then re-emerges.
In addition, NNCPD successfully shows how for the pairs consisting of very similar topics, each topic has different column association as showcased in the $B$ matrix.
Further, it is evident in matrix $C$ that between time slices 10 and 11, this shift in column association happened; thus, it shows the evolution of the column representation of the topics.
\emph{\textbf{Therefore, NNCPD gives additional information that the Direct NMF and Fixed NMF are not able to provide}}.

\subsection{The 20 Newsgroups Dataset Numerical Experiments}
\label{sec:20news}

The 20 Newsgroups dataset is a collection of approximately 20,000 text documents containing the text of messages from 20 different newsgroups on the distributed discussion system Usenet which functioned similarly to current internet discussion forums.  
The documents are partitioned nearly evenly across the 20 newsgroups which can be further classified into six supergroups (computers, for sale, sports/recreation, politics, science, religion) \cite{KL08}. 
We apply NNCPD, Direct NMF, and Fixed NMF to a tensor dataset constructed from this data.   

We construct the tensor using only a subset of the four supergroups: `for sale', `baseball', `atheism' and `space'. 
We randomly select 780 documents for each of `for sale' and `space' newsgroups, and 390 documents for each of `atheism' and `baseball' newsgroups.
We further remove headers, footers and quotes from all of the documents.
We compute the TF-IDF weights across all documents instead of raw word counts to better reflect how important a word is to a document in a collection of documents.
In particular, we apply the TF-IDF vectorizer \cite{TfIdf} from Scikit-Learn \cite{scikit-learn} in Python.
We select a total of 5000 words with the highest TF-IDF weights across all documents, excluding stop-words, words that appear in less than 2 documents, and words that appear in more than 95\% of the documents.

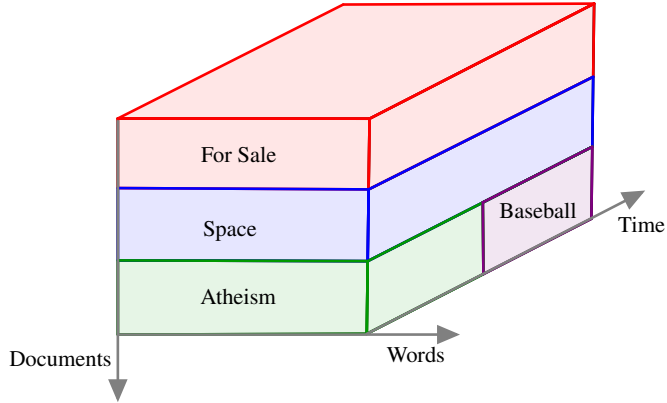
\begin{figure}[h!]
	\centering
	\definecolor{yqqqyq}{rgb}{0.5019607843137255,0.,0.5019607843137255}
\definecolor{qqzzqq}{rgb}{0.,0.6,0.}
\definecolor{qqqqff}{rgb}{0.,0.,1.}
\definecolor{ffqqqq}{rgb}{1.,0.,0.}
\begin{tikzpicture}[line cap=round,line join=round,>=triangle 45,x=10.0cm,y=10.0cm, scale=0.15]
\clip(-3.873122738131945,1.2919101332776169) rectangle (3.030331787420243,5.01216526066976);
\fill[line width=2.pt,color=ffqqqq,fill=ffqqqq,fill opacity=0.10000000149011612] (-0.5318759445019386,3.8520040348560585) -- (-0.5399033278650249,3.2247816209944244) -- (-2.7624945536005,3.236615820270972) -- (-2.7624945536005,3.8538090610622615) -- cycle;
\fill[line width=2.pt,color=ffqqqq,fill=ffqqqq,fill opacity=0.10000000149011612] (-0.5318759445019386,3.8520040348560585) -- (-0.5399033278650249,3.2247816209944244) -- (1.4505099815884113,4.2235887474623475) -- (1.4587618154623903,4.872733012215369) -- cycle;
\fill[line width=2.pt,color=qqqqff,fill=qqqqff,fill opacity=0.10000000149011612] (-0.5399033278650249,3.2247816209944244) -- (-0.5480450149847704,2.588628043025818) -- (-2.7624945536005,2.595059688395816) -- (-2.7624945536005,3.236615820270972) -- cycle;
\fill[line width=2.pt,color=qqqqff,fill=qqqqff,fill opacity=0.10000000149011612] (-0.5399033278650249,3.2247816209944244) -- (-0.5480450149847704,2.588628043025818) -- (1.4426219262469746,3.603061727269339) -- (1.4505099815884113,4.2235887474623475) -- cycle;
\fill[line width=2.pt,color=qqzzqq,fill=qqzzqq,fill opacity=0.10000000149011612] (-0.5480450149847704,2.588628043025818) -- (-0.5563010058611754,1.943543300950237) -- (-2.7624945536005,1.9372616291314155) -- (-2.7624945536005,2.595059688395816) -- cycle;
\fill[line width=2.pt,color=qqzzqq,fill=qqzzqq,fill opacity=0.10000000149011612] (-0.5480450149847704,2.588628043025818) -- (-0.5563010058611754,1.943543300950237) -- (0.4772923055584313,2.4734896534092328) -- (0.4721846564093777,3.109832842759992) -- (1.4426225484257043,3.603110671996069) -- (1.4426219262469746,3.603061727269339) -- cycle;
\fill[line width=2.pt,color=yqqqyq,fill=yqqqyq,fill opacity=0.10000000149011612] (0.4772923055584313,2.4734896534092328) -- (1.4345017213873164,2.964272278309548) -- (1.4426225484257043,3.603110671996069) -- (0.4721846564093777,3.109832842759992) -- cycle;
\fill[line width=2.pt,color=ffqqqq,fill=ffqqqq,fill opacity=0.10000000149011612] (-2.7624945536005,3.8538090610622615) -- (-0.7650801414194005,4.864646314190344) -- (1.4587618154623903,4.872733012215369) -- (-0.5318759445019386,3.8520040348560585) -- cycle;
\draw [line width=1.pt] (-0.5318759445019386,3.8520040348560585)-- (1.4587618154623903,4.872733012215369);
\draw [line width=1.pt] (1.4587618154623903,4.872733012215369)-- (1.4345017213873164,2.964272278309548);
\draw [line width=1.pt] (1.4345017213873164,2.964272278309548)-- (-0.5563010058611754,1.943543300950237);
\draw [line width=1.pt] (-0.5318759445019386,3.8520040348560585)-- (-0.5563010058611754,1.943543300950237);
\draw [line width=1.pt] (-0.5318759445019386,3.8520040348560585)-- (-2.7624945536005,3.8538090610622615);
\draw [line width=1.pt] (-2.7624945536005,3.8538090610622615)-- (-2.7624945536005,1.9372616291314155);
\draw [line width=1.pt] (-0.5563010058611754,1.943543300950237)-- (-2.7624945536005,1.9372616291314155);
\draw [line width=1.pt] (-0.5399033278650249,3.2247816209944244)-- (1.4505099815884113,4.2235887474623475);
\draw [line width=1.pt] (-0.5480450149847704,2.588628043025818)-- (1.4426219262469746,3.603061727269339);
\draw [line width=1.pt] (-0.5480450149847704,2.588628043025818)-- (-2.7624945536005,2.595059688395816);
\draw [line width=1.pt] (0.4721846564093777,3.109832842759992)-- (0.4772923055584313,2.4734896534092328);
\draw [line width=1.pt] (0.4721846564093777,3.109832842759992)-- (1.4426225484257043,3.603110671996069);
\draw [line width=1.pt,color=ffqqqq] (-0.5318759445019386,3.8520040348560585)-- (-0.5399033278650249,3.2247816209944244);
\draw [line width=1.pt,color=ffqqqq] (-0.5399033278650249,3.2247816209944244)-- (-2.7624945536005,3.236615820270972);
\draw [line width=1.pt,color=ffqqqq] (-2.7624945536005,3.236615820270972)-- (-2.7624945536005,3.8538090610622615);
\draw [line width=1.pt,color=ffqqqq] (-2.7624945536005,3.8538090610622615)-- (-0.5318759445019386,3.8520040348560585);
\draw [line width=1.pt,color=ffqqqq] (-0.5318759445019386,3.8520040348560585)-- (-0.5399033278650249,3.2247816209944244);
\draw [line width=1.pt,color=ffqqqq] (-0.5399033278650249,3.2247816209944244)-- (1.4505099815884113,4.2235887474623475);
\draw [line width=1.pt,color=ffqqqq] (1.4505099815884113,4.2235887474623475)-- (1.4587618154623903,4.872733012215369);
\draw [line width=1.pt,color=ffqqqq] (1.4587618154623903,4.872733012215369)-- (-0.5318759445019386,3.8520040348560585);
\draw [line width=1.pt,color=qqqqff] (-0.5399033278650249,3.2247816209944244)-- (-0.5480450149847704,2.588628043025818);
\draw [line width=1.pt,color=qqqqff] (-0.5480450149847704,2.588628043025818)-- (-2.7624945536005,2.595059688395816);
\draw [line width=1.pt,color=qqqqff] (-2.7624945536005,2.595059688395816)-- (-2.7624945536005,3.236615820270972);
\draw [line width=1.pt,color=qqqqff] (-2.7624945536005,3.236615820270972)-- (-0.5399033278650249,3.2247816209944244);
\draw [line width=1.pt,color=qqqqff] (-0.5399033278650249,3.2247816209944244)-- (-0.5480450149847704,2.588628043025818);
\draw [line width=1.pt,color=qqqqff] (-0.5480450149847704,2.588628043025818)-- (1.4426219262469746,3.603061727269339);
\draw [line width=1.pt,color=qqqqff] (1.4426219262469746,3.603061727269339)-- (1.4505099815884113,4.2235887474623475);
\draw [line width=1.pt,color=qqqqff] (1.4505099815884113,4.2235887474623475)-- (-0.5399033278650249,3.2247816209944244);
\draw [line width=1.pt,color=qqzzqq] (-0.5480450149847704,2.588628043025818)-- (-0.5563010058611754,1.943543300950237);
\draw [line width=1.pt,color=qqzzqq] (-0.5563010058611754,1.943543300950237)-- (-2.7624945536005,1.9372616291314155);
\draw [line width=1.pt,color=qqzzqq] (-2.7624945536005,1.9372616291314155)-- (-2.7624945536005,2.595059688395816);
\draw [line width=1.pt,color=qqzzqq] (-2.7624945536005,2.595059688395816)-- (-0.5480450149847704,2.588628043025818);
\draw [line width=1.pt,color=qqzzqq] (-0.5480450149847704,2.588628043025818)-- (-0.5563010058611754,1.943543300950237);
\draw [line width=1.pt,color=qqzzqq] (-0.5563010058611754,1.943543300950237)-- (0.4772923055584313,2.4734896534092328);
\draw [line width=1.pt,color=qqzzqq] (0.4772923055584313,2.4734896534092328)-- (0.4721846564093777,3.109832842759992);
\draw [line width=1.pt,color=qqzzqq] (0.4721846564093777,3.109832842759992)-- (1.4426225484257043,3.603110671996069);
\draw [line width=1.pt,color=qqzzqq] (1.4426225484257043,3.603110671996069)-- (1.4426219262469746,3.603061727269339);
\draw [line width=1.pt,color=qqzzqq] (1.4426219262469746,3.603061727269339)-- (-0.548045014984774,2.588628043025818);
\draw [line width=1.pt,color=yqqqyq] (0.4772923055584313,2.4734896534092328)-- (1.4345017213873164,2.964272278309548);
\draw [line width=1.pt,color=yqqqyq] (1.4345017213873164,2.964272278309548)-- (1.4426225484257043,3.603110671996069);
\draw [line width=1.pt,color=yqqqyq] (1.4426225484257043,3.603110671996069)-- (0.4721846564093777,3.109832842759992);
\draw [line width=1.pt,color=yqqqyq] (0.4721846564093777,3.109832842759992)-- (0.4772923055584313,2.4734896534092328);
\draw [line width=1.pt,color=ffqqqq] (-2.7624945536005,3.8538090610622615)-- (-0.7650801414194005,4.864646314190344);
\draw [line width=1.pt,color=ffqqqq] (-0.7650801414194005,4.864646314190344)-- (1.4587618154623903,4.872733012215369);
\draw [line width=1.pt,color=ffqqqq] (1.4587618154623903,4.872733012215369)-- (-0.5318759445019386,3.8520040348560585);
\draw [line width=1.pt,color=ffqqqq] (-0.5318759445019386,3.8520040348560585)-- (-2.7624945536005,3.8538090610622615);
\draw [->,line width=1.pt,color=gray]  (-2.7624945536005,3.8301320714646295) -- (-2.7624945536005003,1.3372616291314153);
\draw [->,line width=1.pt, color=gray] (-2.7624945536005003,1.9372616291314153) -- (0.27001720578376043,1.9372616291314155);
\draw [->,line width=1.pt, color=gray] (-0.5563010058611753,1.9435433009502372) -- (1.9116169048637734,3.214959917085313);
\draw (-3.7954244501780432,1.860311412939111) node[anchor=north west] {$\text{Documents}$};
\draw (-0.4496989184434374,1.901955213331613) node[anchor=north west] {$\text{Words}$};
\draw (1.6013627562131313,3.0487865210352635) node[anchor=north west] {$\text{Time}$};
\draw (0.5481787109849456,3.1730466151103374) node[anchor=north west] {$\text{Baseball}$};
\draw (-2.1023449392481787,2.412897000758019) node[anchor=north west] {$\text{Atheism}$};
\draw (-2.094258241223154,3.6744218926618666) node[anchor=north west] {$\text{For Sale}$};
\draw (-2.0780848451731046,3.0193993526348686) node[anchor=north west] {$\text{Space}$};
\end{tikzpicture}
	\caption{Visualization of the construction of the 20 Newsgroups tensor.}
	\label{fig:20newsstructure3}
\end{figure}

We build the three-mode tensor of size $234\times 5000\times 10$ such that the first mode represents documents, the second represents words, and the third represents time. 
The structure of the tensor is shown in Figure \ref{fig:20newsstructure3}. 
The top red part consists of 780 `for sale' documents evenly spread across 10 time slices. 
Similarly, the blue part consists of 780 `space' documents evenly spread across 10 time slices.
Lastly, the green part consists of 390 `atheism' documents evenly spread across time slices 1-5,
and the purple part consists of 390 `baseball' documents evenly spread across time slices 6-10.

\begin{figure}[h!]
	\centering
	\includegraphics[height=2.1in]{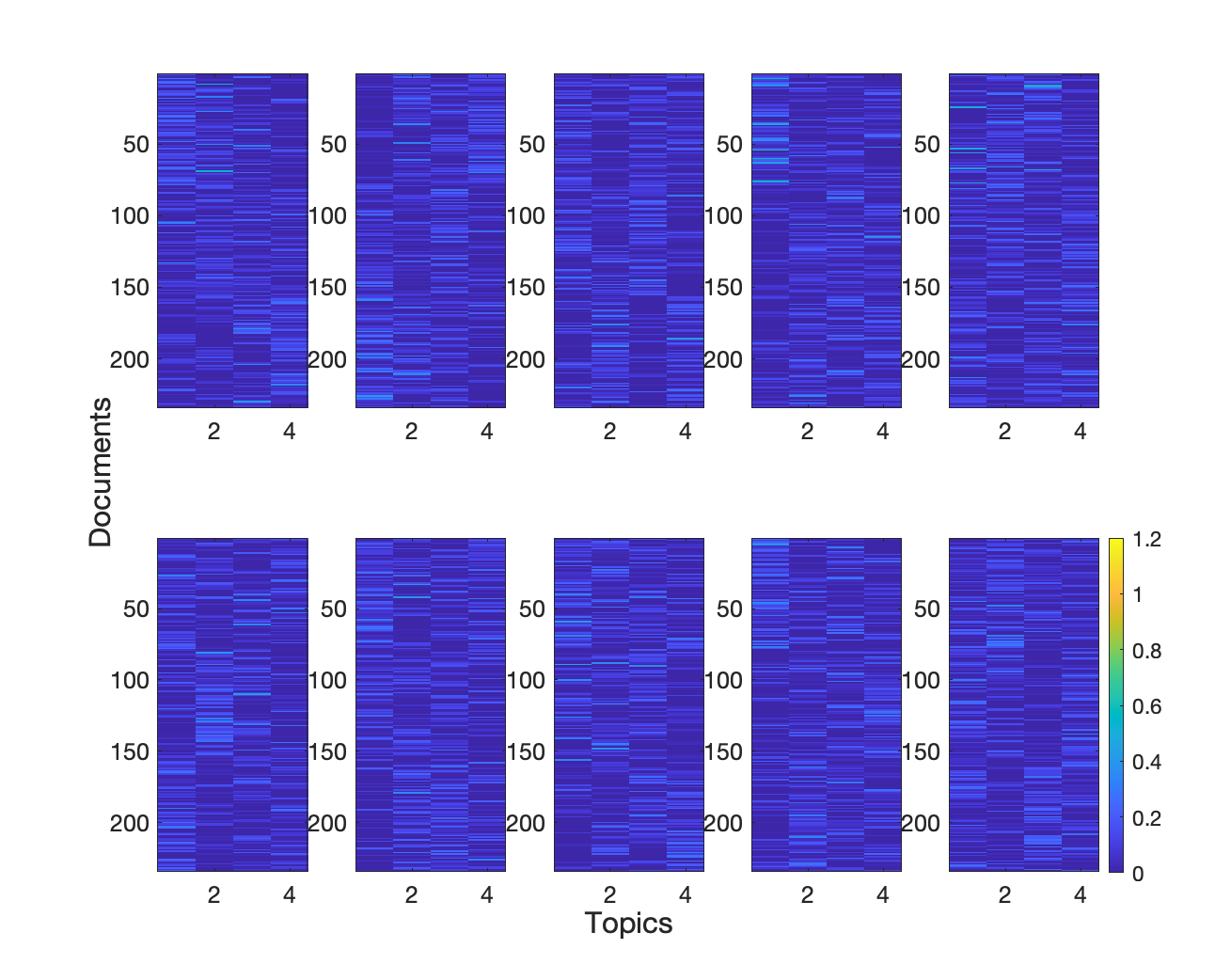}
	\includegraphics[height=2.1in]{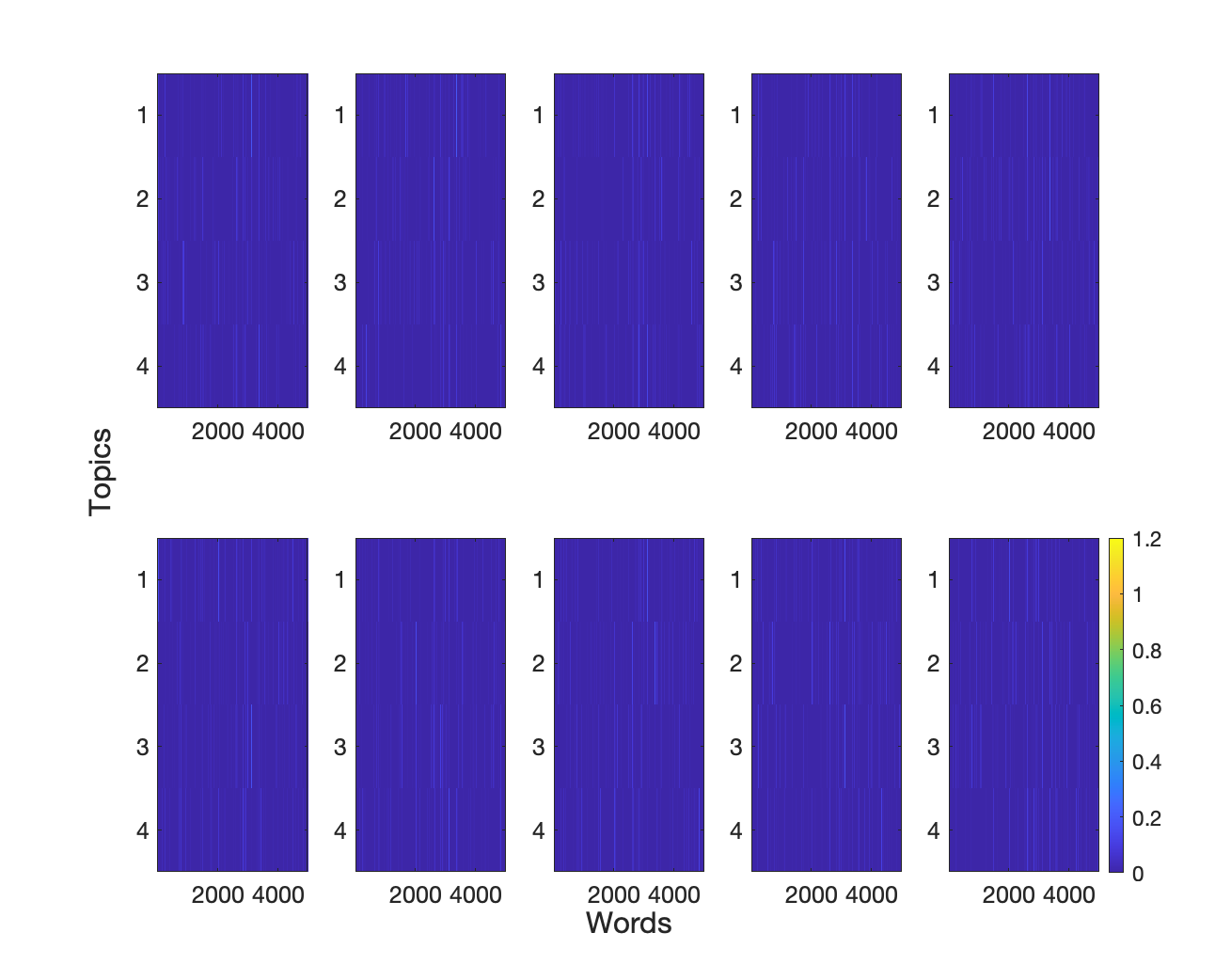} 
	\caption{Direct NMF performed slice by slice for tensor visualized in Figure \ref{fig:20newsstructure3} with rank 4. Left: Direct NMF $A$ factors showcasing the document-representation of topics. Right: Direct NMF $S$ factors showcasing the word-representation of topics aligned with the $A$ factors. 
	Document variation is not clear in the $A$ factors.
    Further, temporal information is difficult to identify in the $A$ and $S$ factor matrices. The Direct NMF reconstruction error is 46.717.}\label{HWSlices7-13-2020}
\end{figure}

\begin{figure}[h!]
	\centering
	\includegraphics[height=1.25in]{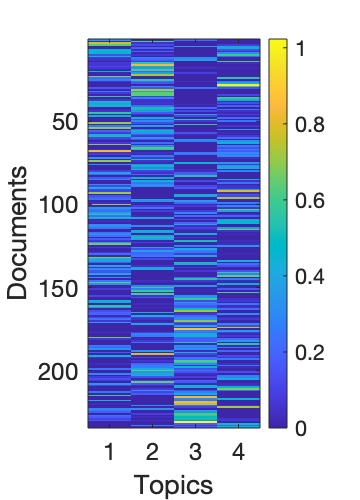}
	\includegraphics[height=1.25in]{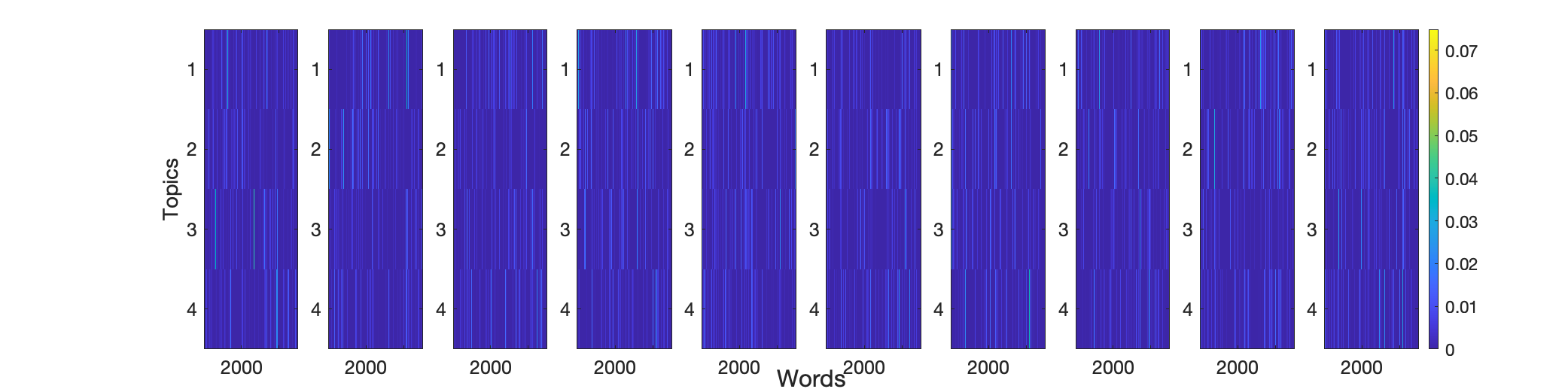}
	\caption{Fixed NMF performed on matricized tensor along the first mode for tensor visualized in Figure \ref{fig:20newsstructure3} for rank 4. Left: Fixed NMF common $A$ factor for each slice showcasing the document-representation of topics. Right: Fixed NMF $S$ factors for each slice showcasing the word-representation of topics. 
	Document variation is not clear in the $A$ factor.
    Further, temporal information is difficult to identify in the $S$ factor matrices due to the dimensionality of the matrices.
	Fixed NMF reconstruction error is 46.8953.} \label{SSlicesFixA7-13-2020} 
\end{figure}

\begin{figure}[h!]
	\centering
	\includegraphics[height=1.25in]{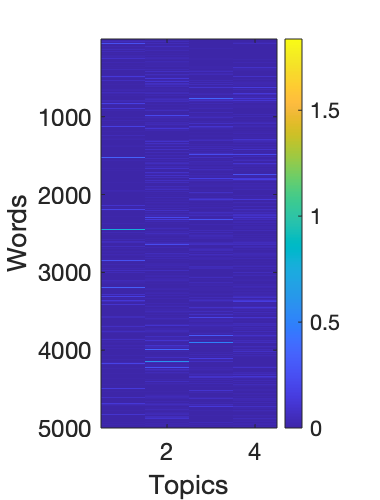}
	\includegraphics[height=1.25in]{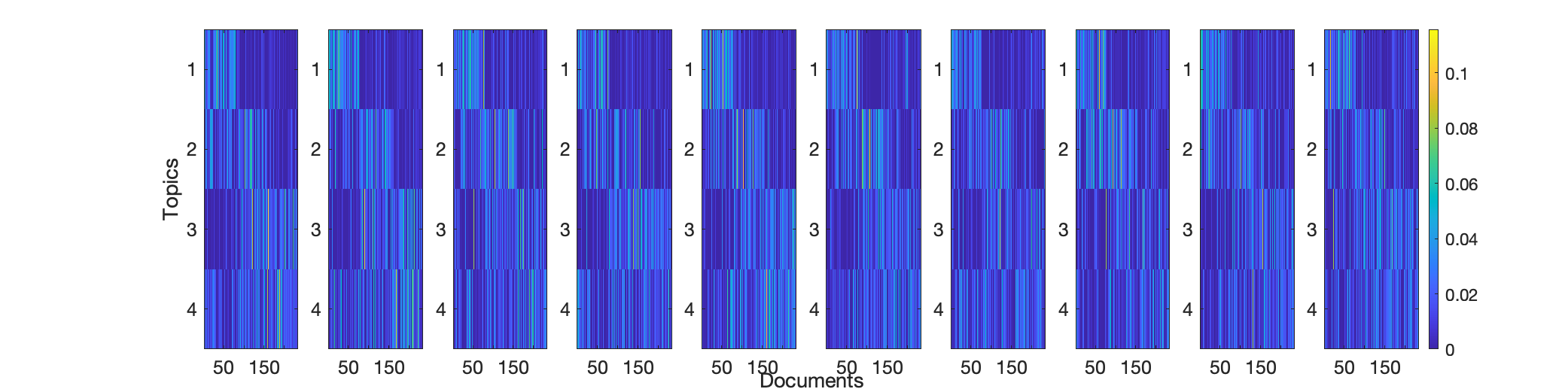}
	\caption{Fixed NMF performed on matricized tensor along second mode for tensor visualized in Figure \ref{fig:20newsstructure3} for rank 4. Left: Fixed NMF common $A$ factor for each slice showcasing the word-representation of topics. Right: Fixed NMF $S$ factors showcasing the document-representation of topics for each slice. 
	Document variation is not clear for all topics in the $S$ factors.
    Further, temporal information is difficult to identify in the $S$ factor matrices due to the dimensionality of the matrices.
	Fixed NMF reconstruction error is 46.99.} \label{SSlicesFixA(transposed)7-13-2020} 
\end{figure}

\begin{figure}[h!]
	\centering
	\includegraphics[height=1.25in]{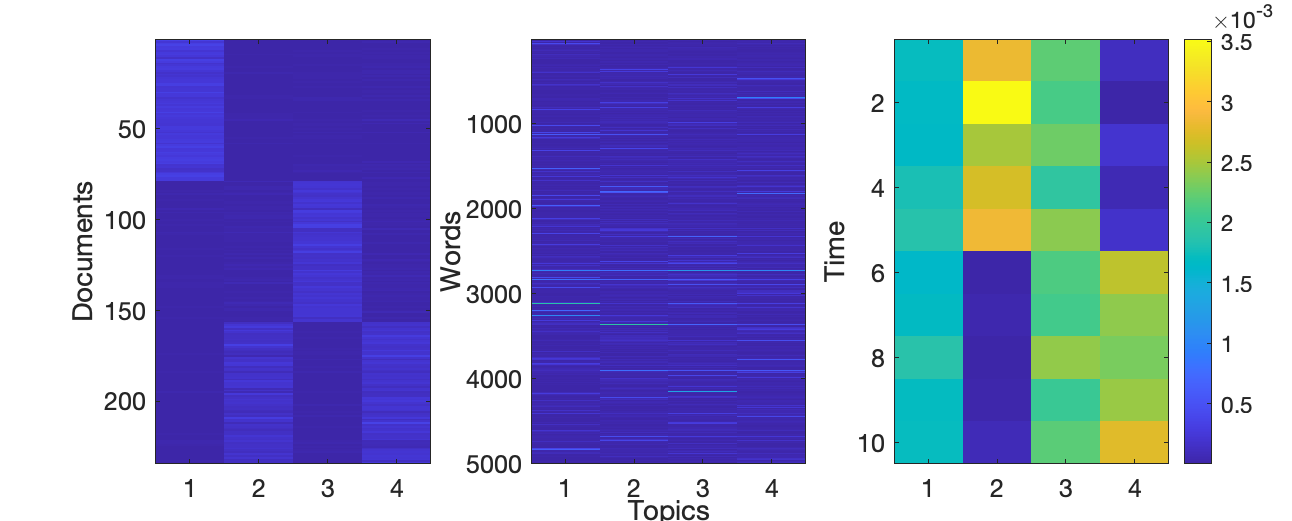}
	\caption{NNCPD factors $A$, $B$, and $C$ (Left, Center, and Right, respectively) for tensor visualized in Figure \ref{fig:20newsstructure3}.  We are able to identify groups of documents sharing topics from the factor $A$, and temporal information from factor $C$. We see that topics one and three (`for sale' and `space' according to Table \ref{table:20news3}) persist through all time while topic 2 (atheism) transitions into topic 4 (baseball) at time slice six.  NNCPD reconstruction error is 47.0149}\label{ABCwith7-11-2020}
\end{figure}

We perform Direct NMF with rank 4 in Figure \ref{HWSlices7-13-2020} and Fixed NMF with rank 4 in Figures \ref{SSlicesFixA7-13-2020} and  \ref{SSlicesFixA(transposed)7-13-2020} by matricizing along modes 1 and 2 respectively. 
We compute NNCPD for the entire tensor; the NNCPD factor matrices $A$, $B$, and $C$ are presented in Figure \ref{ABCwith7-11-2020}.  
Further, we present in Table \ref{table:20news3} the top 10 keywords for each of the topics obtained from the $B$ matrix.
We present the associated reconstruction errors in the captions of the figures.


\begin{table}
	\centering
	\begin{tabular}{|p{1.5cm}|p{1.5cm}|p{1.5cm}|p{1.5cm}|} \hline
		\textbf{Topic 1} & \textbf{Topic 2} & \textbf{Topic 3} & \textbf{Topic 4} \\ \hline
		sale & god & space & team \\ \hline
		00 & people & nasa & game \\ \hline
		offer & don & shuttle & year \\ \hline
		new & just & launch & baseball \\ \hline
		shipping & religion & orbit & games \\ \hline 
		drive & think & like & hit \\ \hline
		condition & does & earth & think \\ \hline
		email & believe & just & runs \\ \hline
		sell & say & moon & don \\ \hline
		asking & atheists & program & braves \\ \hline
	\end{tabular}
	\caption{Topic keywords from NNCPD $B$ factor matrix for the 20 Newsgroups dataset.}\label{table:20news3}
\end{table}

Figure~\ref{ABCwith7-11-2020} displays the three factor matrices obtained by performing NNCPD with rank 4 on the 3-mode newsgroups tensor. 
The matrix $B\in \mathbb{R}^{5000 \times 4}_{\ge 0}$ showcases word-representation of topics i.e., the words that are associated to each topic.
We refer to the columns of $B$ as topics.
In Table~\ref{table:20news3}, we display the top 10 words with the highest magnitude in each of the columns of $B$; thus obtaining top-keyword representation of these topics.
We observe that each of these keywords list is cohesive and represents a meaningful topic associated with one of the supergroups: `for sale', `atheism', `space', and `baseball'.
Furthermore, we learn about the variation of these topics across time and documents in the remaining factor matrices as follows.
In Figure~\ref{fig:20newsstructure3}, the documents in the `for sale' category occupy the first 78 document slots throughout the entire 10 time slices.
We observe these two facts in the first columns of matrices $A \in \mathbb{R}^{234 \times 4}_{\ge 0}$ and $C \in \mathbb{R}^{10 \times 4}_{\ge 0}$. 
Similarly, the documents in the `space' category occupy the second 78 document slots throughout the entire 10 time slices.
We observe these two facts as well in the third columns of matrices $A$ and $C$.
Furthermore, we observe that the documents in the `atheism' and `baseball' categories occupy the last 78 document slots. 
The `atheism' documents are present in time slices 1 through 5, and the baseball documents from time slice 6 to 10.
We observe these facts in the second and fourth columns of matrices $A$ and $C$.

\textbf{\emph{Indeed, we see that not only are the keywords in Table \ref{table:20news3} meaningfully associated to the latent topics in the dataset, but also the NNCPD $A$ factor in Figure \ref{ABCwith7-11-2020} captures the topic variation across documents and the NNCPD $C$ factor exhibits the temporal topic information, while this information is difficult to glean from Direct NMF and Fixed NMF due to the dimensionality of the factor matrices.}}

\subsection{Noise  Dataset Robustness Numerical Experiments}
\label{sec: robustness}
In this section, we perform numerical experiments on noise datasets to test robustness of NNCPD  compared to Direct NMF and Fixed NMF. 

\subsubsection{Construction of the Noise  Dataset}

Given a nonnegative deterministic tensor $X \in \R_{\geq0}^{n_1 \times n_2 \times n_3}$ of the form $ X = \sum\limits_{i=1}^{r^*} a_i \otimes b_i \otimes c_i,$ 
with given vectors $a_i$, $b_i$ and $c_i$, and 
$r^*$ denoting the exact rank of $X$, we define a tensor $T$ with noisy measurements,
\begin{equation}T  = X + N,
\end{equation} where  $N $ is a noise tensor.
We will 
implement
two   types of noise tensor $N$ in the experiments below.
First, we consider
a noise tensor $N:=|Z| $, for 
 $Z \in \R  ^{n_1 \times n_2 \times n_3}$  defined as 
\begin{equation}\label{eq:noise}
    Z = \sigma \sum\limits_{i=1}^{r_N} n_{ai} \otimes n_{bi} \otimes n_{ci},
\end{equation}  
where $r_N$ denotes the exact rank of $Z$, $\sigma >0$ is a noise parameter, and the vectors $n_{ai} \in R^{n_1}$, $n_{bi} \in R^{n_2}$, and $n_{ci} \in R^{n_3}$ have entries sampled from $\mathcal N (0,1)$, the standard normal distribution. Second, we consider a noise tensor $N=\sigma |\tilde Z|$, where the entries of the tensor $\tilde Z$ are sampled from the standard normal distribution. 

We carry out the robustness experiments by modulating the variance of the noise parameter $\sigma$ and rank of the noise tensor $r_N$, and examine the resulting reconstruction error. 
Here the  NNCPD reconstruction error is $\| \hat T - X\|_F$,
where   $\hat T$ denotes the NNCPD reconstruction of the tensor $T$, 
as we do not wish to fit the noise $N$.  
The reconstruction error is defined similarly for Direct NMF and Fixed NMF, which is $
\| \tilde T - X\|_F$ where $\tilde T$ is the Direct NMF or Fixed NMF reconstruction of the tensor $T$.
\begin{figure} [h!]
\centering
\includegraphics[scale = 0.18]{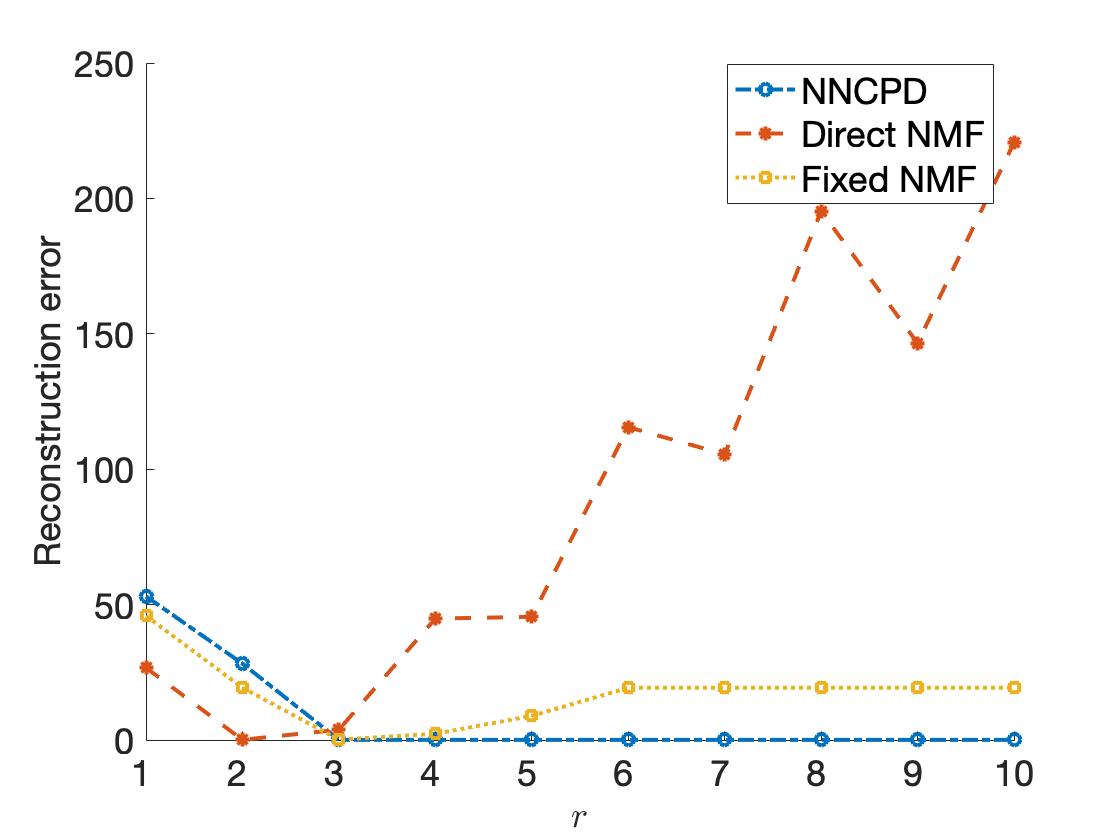}
\hspace{1cm}
\includegraphics[scale = 0.18]{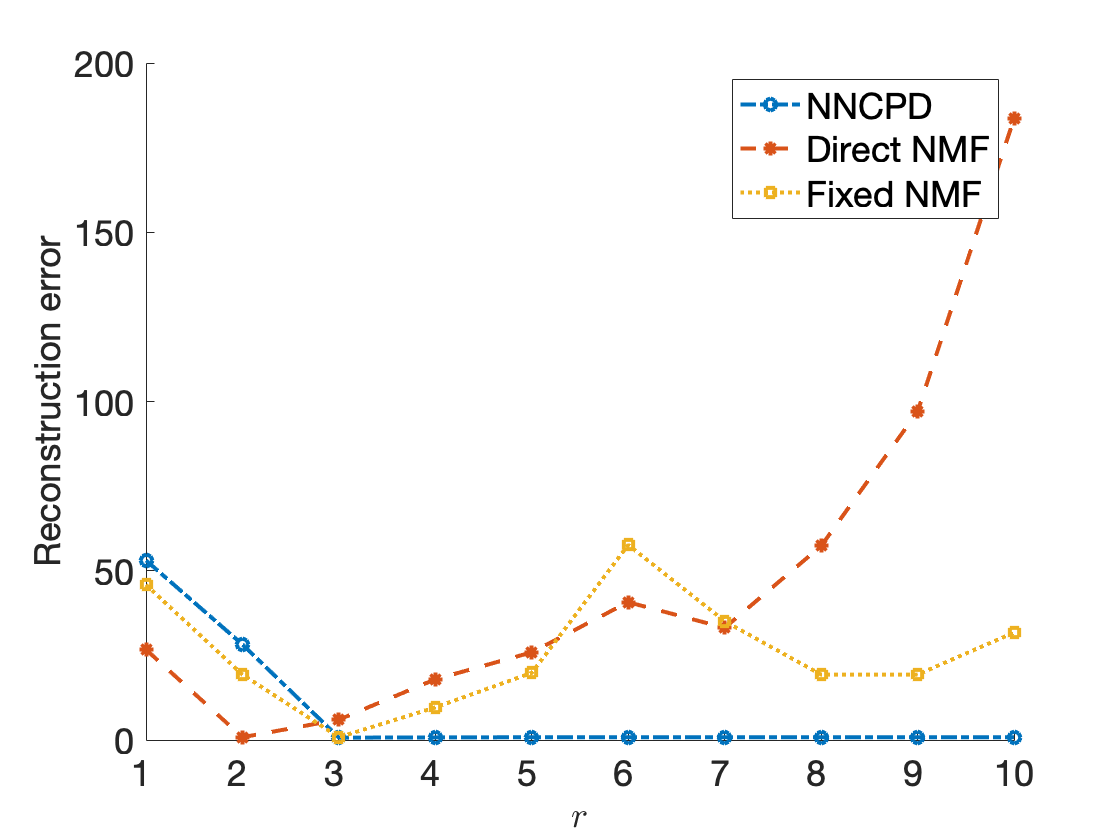}
\\
\includegraphics[scale = 0.18]{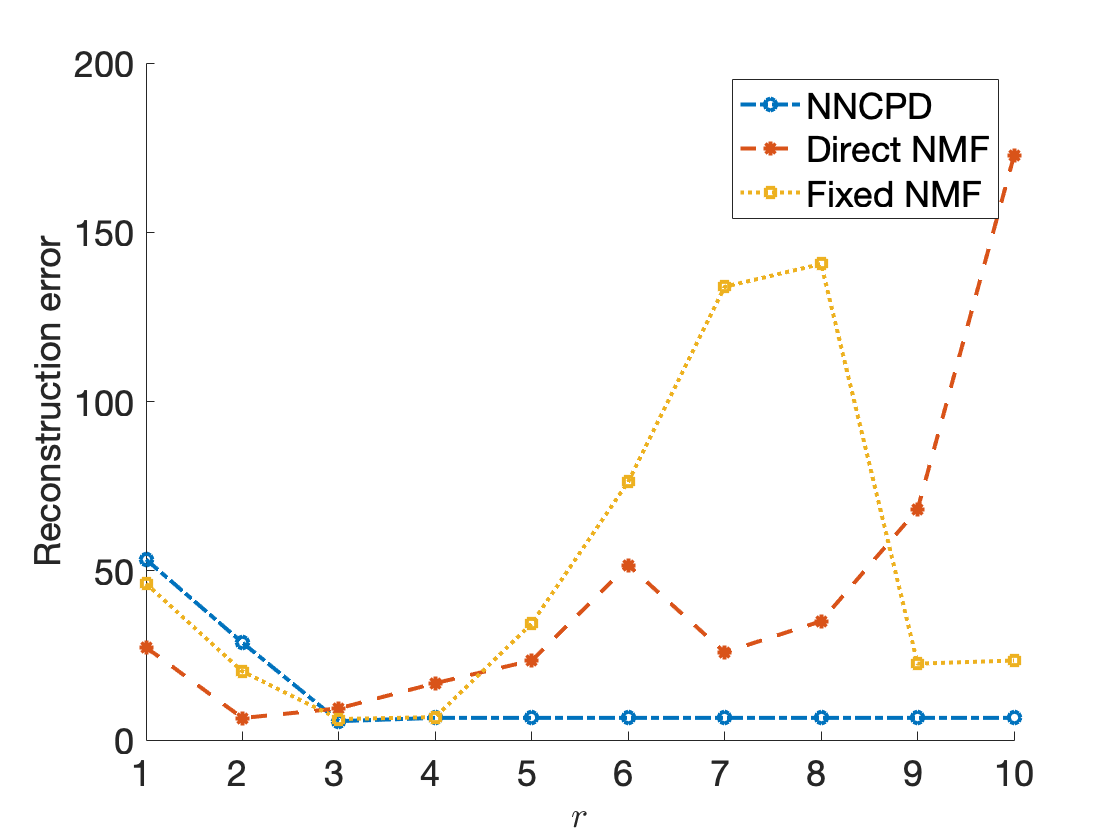}
\hspace{1cm}
\includegraphics[scale = 0.18]{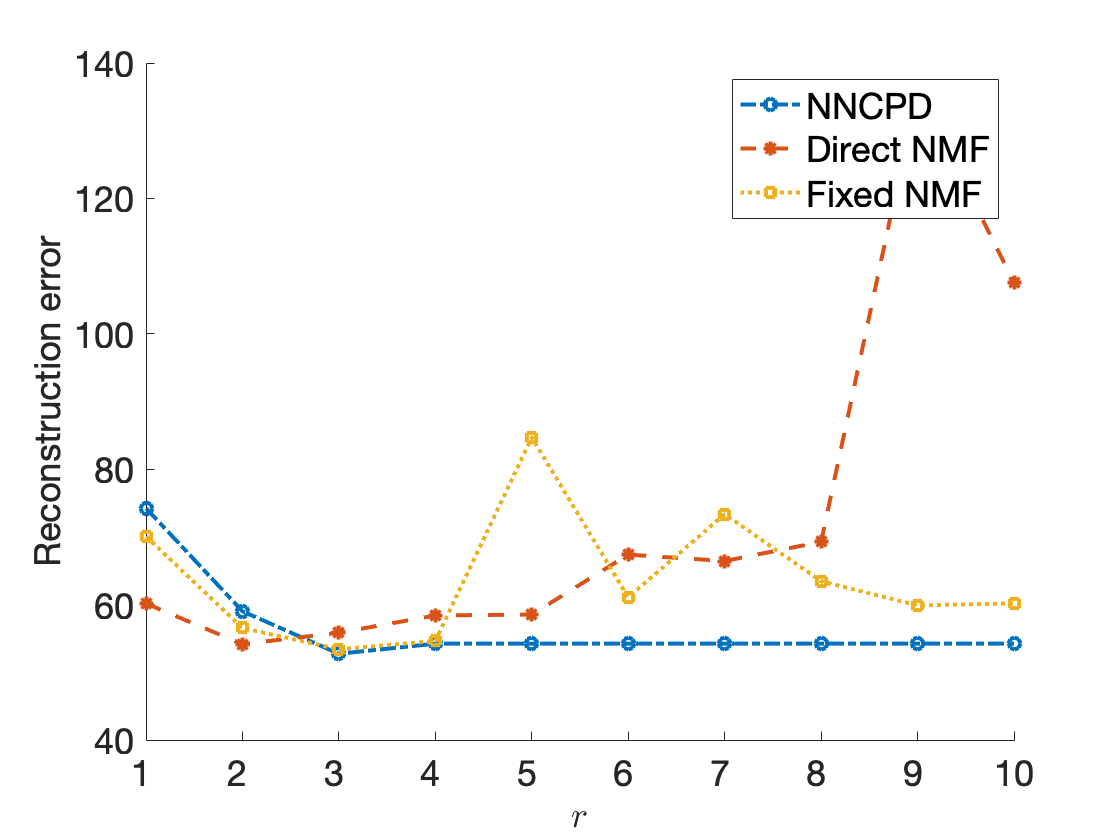}
\caption{In these experiments, we add to all of the entries of the tensor 
{$X$} positive noise drawn from the standard normal distribution.  The noise 
is added in the form of a rank 1 tensor $N$. We report the median of the reconstruction error of 
{NNCPD, Direct NMF, and Fixed NMF}
over 50 runs.
We let the noise parameter vary $\sigma = 10^{-3}, 10^{-2}, 10^{-1}, 1$ from left to right (starting from the top left plot). 
}
\label{fig1:rank1_full_noise}
\end{figure}

\subsubsection{Experiment Output on Noise Dataset}
\label{sec:noisedata}

\begin{figure} [h!]
\centering
\includegraphics[scale = 0.18]{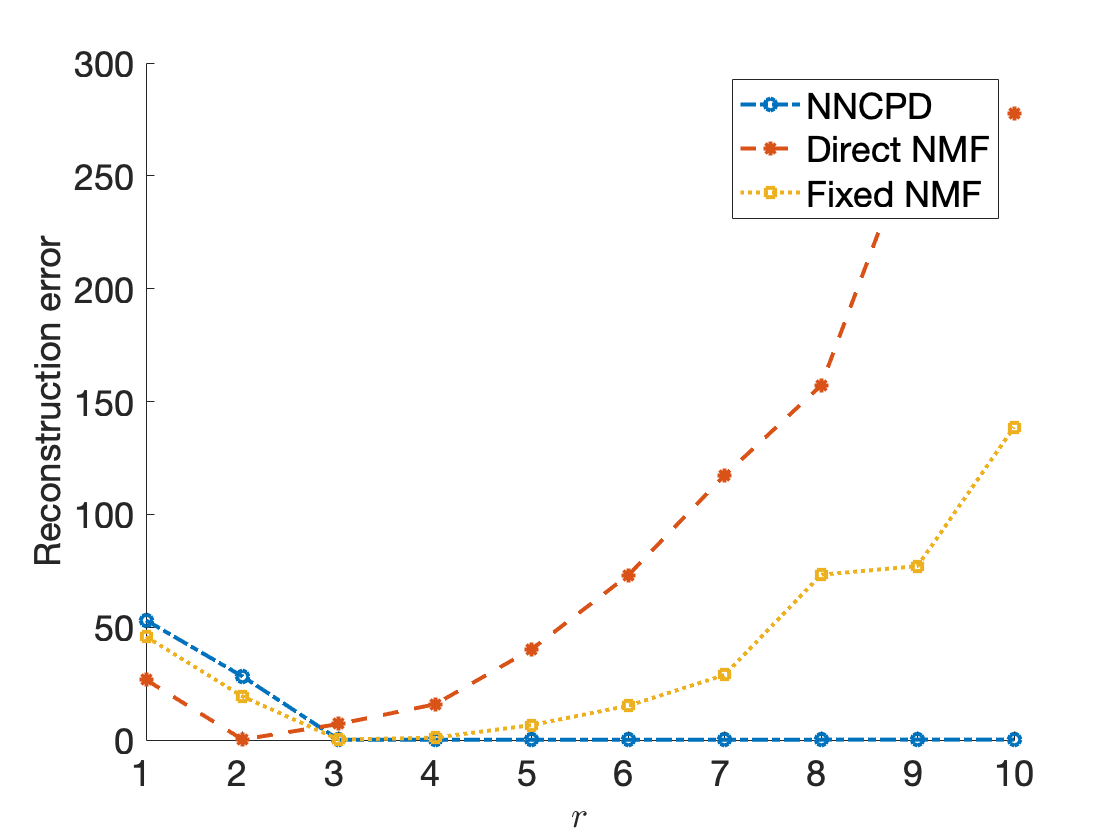}
\hspace{1cm}
\includegraphics[scale = 0.18]{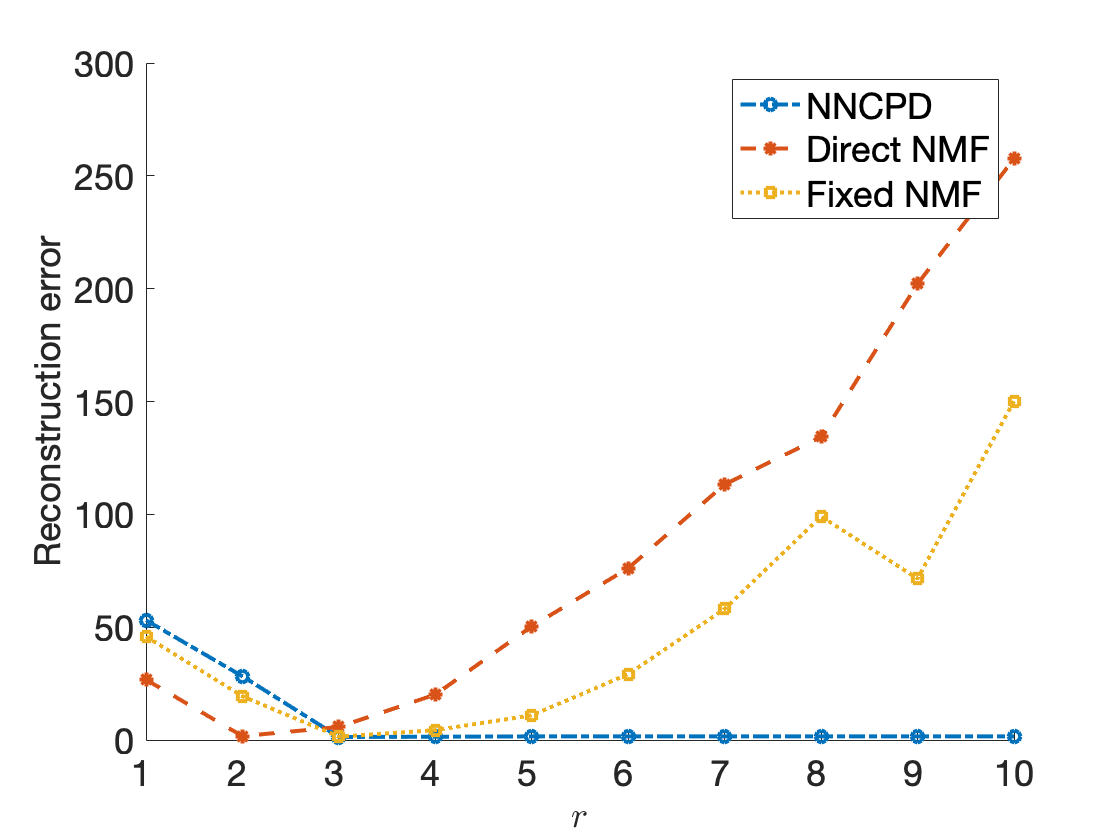}
\\
\includegraphics[scale = 0.18]{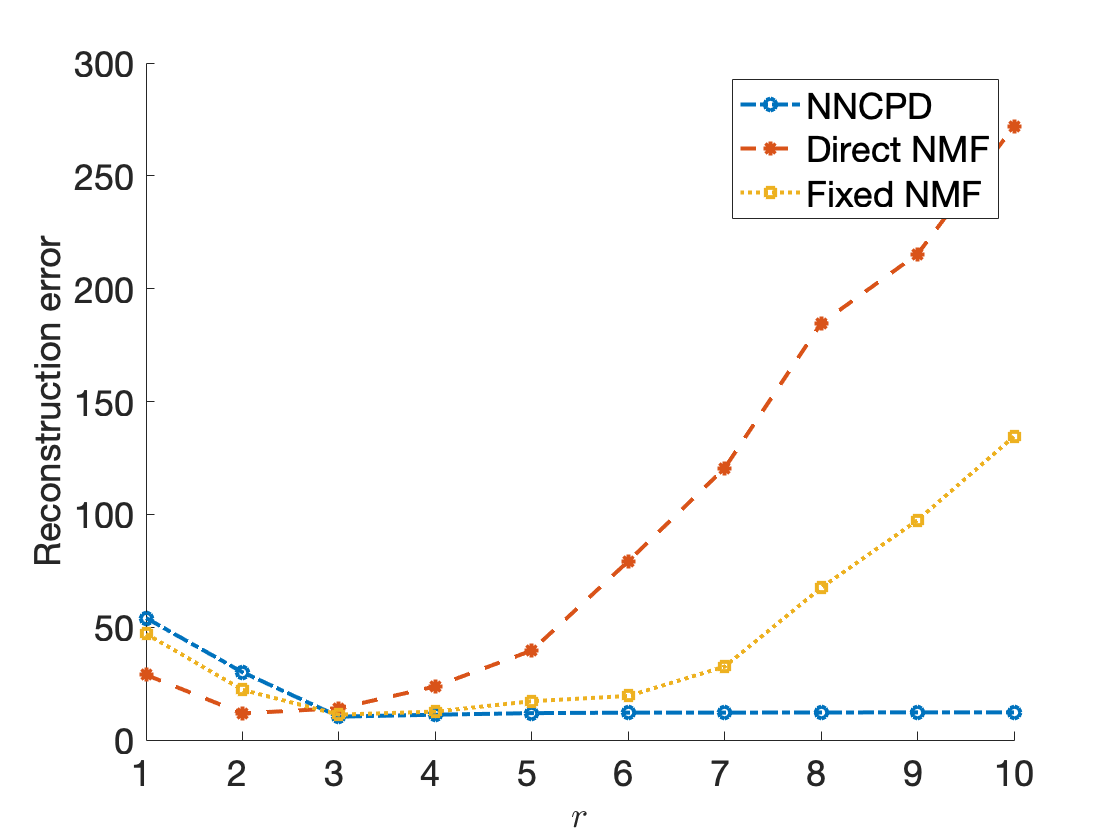}
\hspace{1cm}
\includegraphics[scale = 0.18]{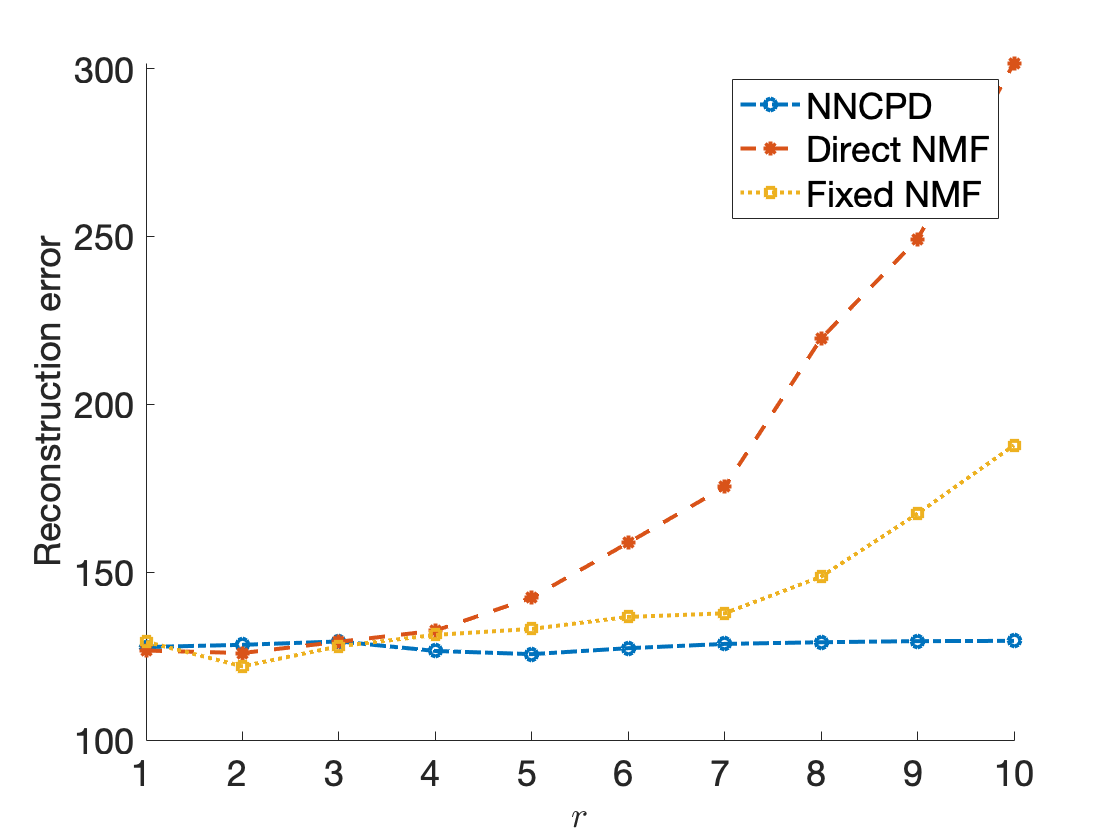}
\caption{In these experiments, we add to all of the entries of the tensor $X$ positive noise drawn from the standard normal distribution. The noise is added in the form of a rank 2 tensor $N$. We report the median of the reconstruction error of NNCPD, Direct NMF, and Fixed NMF over 50 runs.
We let the noise parameter vary $\sigma = 10^{-3}, 10^{-2}, 10^{-1}, 1$ from left to right (starting from the top left plot).}
\label{fig1:rank2_full_noise}
\end{figure}

\begin{figure} [h!]
\centering
\includegraphics[scale = 0.18]{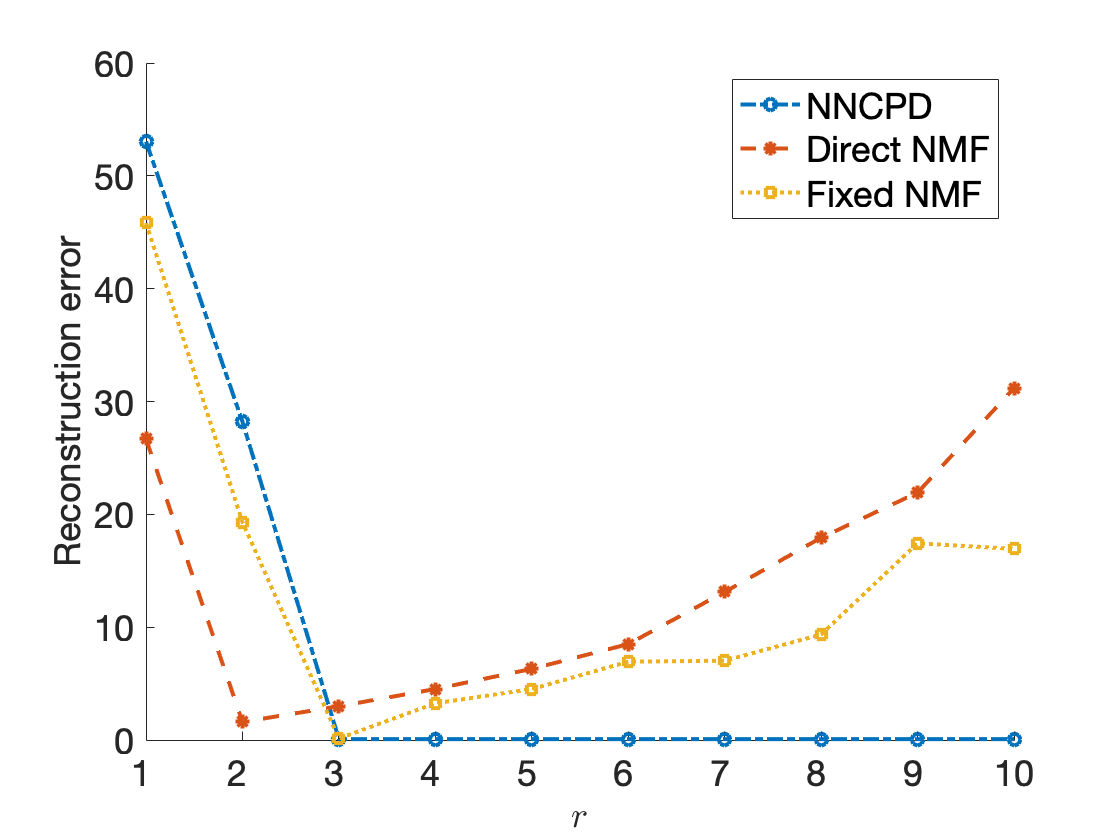}
\hspace{1cm}
\includegraphics[scale = 0.18]{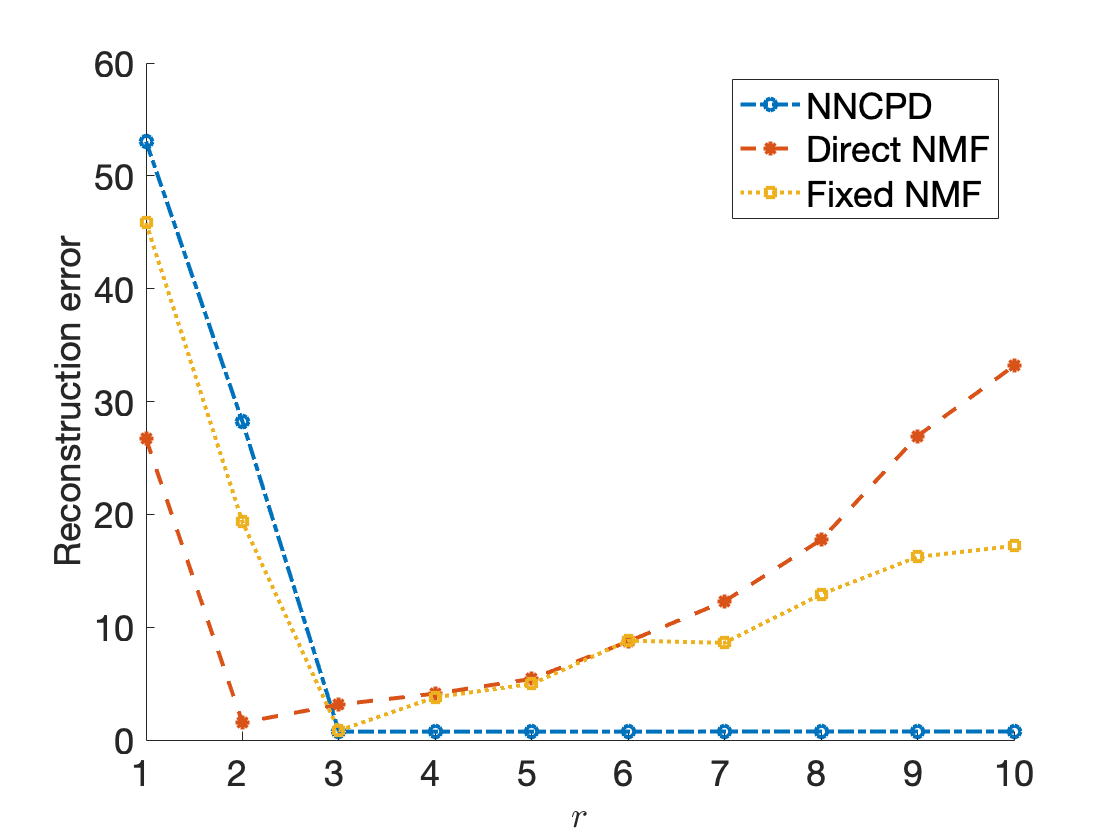}
\\
\includegraphics[scale = 0.18]{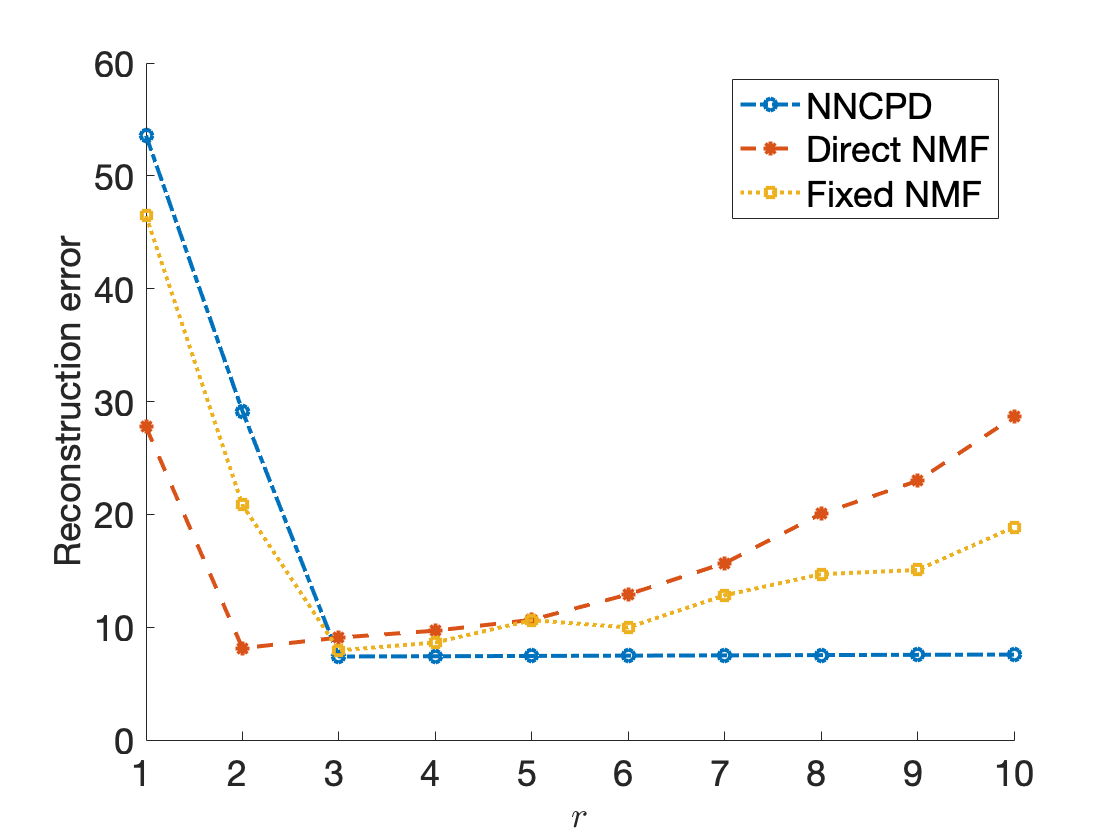}
\hspace{1cm}
\includegraphics[scale = 0.18]{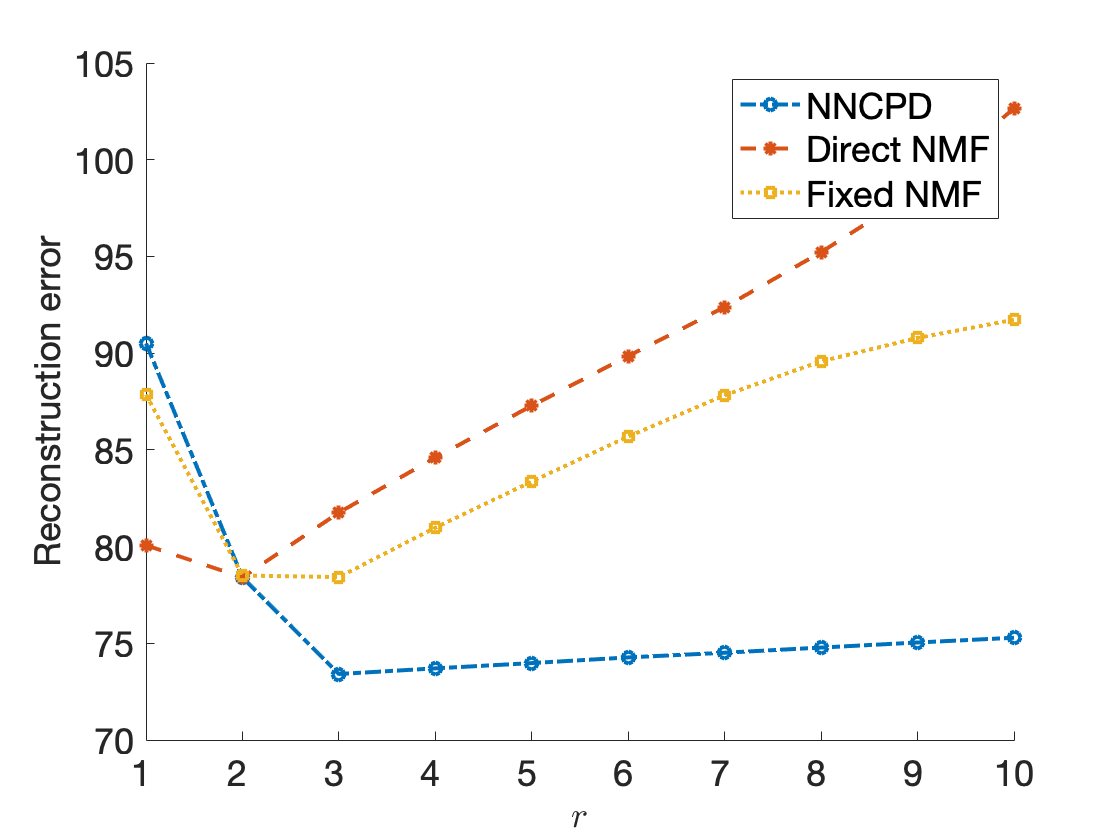}
\caption{In these experiments, we add to all of the entries of the tensor $X$ positive noise drawn from the standard normal distribution $Z^\prime$. We report the median of the reconstruction error of NNCDP, Direct NMF,and Fixed NMF over 50 runs.
We let the noise parameter vary $\sigma = 10^{-3}, 10^{-2}, 10^{-1}, 1$ from left to right (starting from the top left plot)}
\label{fig1:full_noise}
\end{figure}

We experiment to test robustness of NNCPD by adding noise to the tensor $X$ described in the first experiment in Sect.~\ref{ss: sinusoidal data} (see Figure~\ref{fig1:T_exp1}).
The tensor models various dynamic situations where a topic emerges, and evolves.
In Section~\ref{ss: sinusoidal data}, we studied the interpretability of the dynamic topics obtained when applying NNCPD to the tensor.
The Frobenius norm of $X$ is $\|X\|_F \approx 117.1778$, and the exact rank of $X$ is $r^* =3$.
We compute  reconstruction errors  using Direct NMF, Fixed NMF and NNCPD for various ranks $r$. We report the median\footnote{We choose not to report the mean because Direct NMF and Fixed NMF are often not stable or robust, and therefore result in arbitrarily large values making it hard to observe the behavior of NNCPD as the rank varies.} of the reconstruction error over 50 runs in Figures \ref{fig1:rank1_full_noise}-\ref{fig1:full_noise}. We let the noise parameter vary, $\sigma = 10^{-3}, 10^{-2}, 10^{-1}, 1$ from left to right (starting from the top left plot).

Results in Figure~\ref{fig1:rank1_full_noise} are for a noise tensor $N$ with rank one ($r_N = 1$). We compute the norm of $N$ for each noise parameter $\sigma$,
\[ \|N \|_F \approx \begin{cases} 
      0.0531 & \mbox { for } \sigma = 10^{-3}, \\
      0.8174 & \mbox { for } \sigma = 10^{-2}, \\
      6.5349 & \mbox { for } \sigma = 10^{-1}, \\ 
      54.2263 & \mbox { for } \sigma = 1.
   \end{cases}.
\]
For large values of $\|N\|_F$ and while $\|N\|_F < \|X\|_F$, we notice that 
{the NNCPD reconstruction error}  $\| \hat T - X \|_F \approx \|N\|_F$ for $r \geq r^* + r_N$ (in this case $r \geq 4$), and
remains  stable. 

Results in Figure~\ref{fig1:rank2_full_noise} are for a noise tensor $N$ with rank two ($r_N = 2$). We compute the norm of $N$ for each noise parameter $\sigma$,

\[ \|N \|_F \approx \begin{cases} 
      0.1554 & \mbox { for } \sigma = 10^{-3}, \\
      1.6394 & \mbox { for } \sigma = 10^{-2}, \\
      12.3065 & \mbox { for } \sigma = 10^{-1}, \\ 
      129.9499 & \mbox { for } \sigma = 1.
   \end{cases}.
\]
For large values of $\|N\|_F$ and while $\|N\|_F < \|X\|_F$, we notice that NNCPD reconstruction error $\| \hat T -X \|_F \approx \|N\|_F$ for $r \geq r^* + r_N$ (in this case $r \geq 5$) and remains  stable. For $\sigma = 1$, i.e., when $\|N\|_F = 129.9499 > 117.1778 = \|X\|_F$, we see that NNCPD does not detect that $r=3$ is the rank of the true tensor $A$. At the same time, the reconstruction error is minimum for $r=5$. This suggests that since $\|N\|_F > \|X\|_F$,  NNCPD is fitting to the noise tensor $N$ and not the true tensor $X$ because there is better hope minimizing the error fitting to the noise tensor.

Results in Figure~\ref{fig1:full_noise} are for a noise tensor $N = \sigma |\tilde {Z} |$, where the entries of the tensor $\tilde {Z} \in \R^{n_1 \times n_2 \times n_3}$ are sampled from the standard normal distribution.

\[ \|N \|_F \approx \begin{cases} 
      0.0911 & \mbox { for } \sigma = 10^{-3}, \\
      0.9226 & \mbox { for } \sigma = 10^{-2}, \\
      9.2329 & \mbox { for } \sigma = 10^{-1}, \\ 
      91.9537 & \mbox { for } \sigma = 1.
   \end{cases}.
\]
For these experiments, due to the distribution of its entries, the tensor $N$ likely has much higher CP rank than the tensor $X$.
We notice that NNCPD reconstruction error $\| \hat T - X \|_F < \|N\|_F$ for $r < 10$; in contrast, the reconstruction error of Fixed NMF and Direct NMF often exceed $\|N\|_F$ for $r < 10$.
The experiments suggest that the NNCPD reconstruction for overestimates of the rank can tolerate more noise when the noise tensor is of high rank than of low rank (with the same magnitude).

\textbf{\emph{In conclusion, the numerical experiments for robustness of 
NNCPD suggest that this decomposition is stable and robust to noise.
In contrast, Direct NMF and Fixed NMF display unstable behaviour as we overestimate the rank of the tensor. 
Further, the experiments show that these kinds of tests 
can potentially estimate the rank (or an upper bound on the number of topics) of the tensor using NNCPD. We also notice a stable behaviour for NNCPD with a cusp or a minimum at $r^*$ for tolerable noise.}} 

\section{Conclusion}\label{sec:conclusion}

In this article, we have proposed a new methodology for large-scale dynamic topic modeling arising from the explosion of data in the information era.  
Previous works primarily employ NMF-based methods to decompose high dimensional data tensors that have one or more temporal dimensions. These methods seek to factorize a matricized version of the tensor sliced along the temporal  dimension. Often a tensor is broken directly into time slices and each slice is decomposed individually using NMF (Direct NMF). Alternatively,  authors in \cite{cichocki2007nonnegative} decompose the concatenated time slices of a tensor with one of the factors being fixed (Fixed NMF). There is a significant disadvantage of such NMF-based methods, in that the temporal mode of the data is not respected, thereby neglecting or oversimplifying the temporal information. To address this issue, we proposed using   the method of nonnegative CP tensor decompostion (NNCPD)  where the tensor is directly decomposed into a minimal sum of outer products of nonnegative vectors. In this way, critical temporal information is preserved, and events such as topic evolution, emergence and fading become significantly easier to identify.  

In order to compare NMF methods with our NNCPD approach, we performed numerical experiments applied to deterministic synthetic datasets  (Sect.~\ref{sec:syntheticdataset}),  a real-life news dataset (Sect. \ref{sec:20news}), and   on synthetic noisy tensor datasets (Sect.~\ref{sec: robustness}).
 We demonstrated how the factors of NNCPD can be interpreted for dynamic topic modeling. For example, we observe for 3-mode tensors that the three NNCPD factor matrices display the topic representation for each word, topic representation for each document, and the evolution of the topics through time, respectively. The  results of NNCPD exhibit significant advantages when compared to results of Direct NMF and Fixed NMF. In particular, 
NNCPD succeeds in all the experiments in detecting topic evolution and emergence that the Direct NMF and Fixed NMF failed to discover. 
In addition, 
for the real-life 20 newsgroup dataset (Sect. \ref{sec:20news}),
we find that by 
applying NNCPD, we can clearly identify times of topic emergence and fading using the temporal factor matrix, and can clearly determine the document structure of the tensor using the document factor matrix (Figure \ref{ABCwith7-11-2020}). 

Finally, for noisy tensor datasets
(Sect.~\ref{sec: robustness})
the experiment output suggests that NNCPD is superior to Direct NMF and Fixed NMF in stability and robustness to noise.
Further, we observe that NNCPD is stable to overestimates of the rank, suggesting that the NNCPD-rank of a tensor can be estimated by producing a plot showcasing the reconstruction error of NNCPD with various ranks. 

In summary, NNCPD proves to be a powerful  tool for  dynamic topic modeling, and compares favorably with typical   NMF-based methods. We believe that our work provides an introduction to and evidence for the value of further exploration of this new approach of dynamic topic modeling through NNCPD.

\bibliographystyle{alpha}
\bibliography{mybib}

\end{document}